\documentclass[lettersize,article,twoside]{IEEEtran}
\usepackage{amsmath,amsfonts}
\usepackage{algorithmic}
\usepackage{algorithm}
\usepackage{array}
\usepackage[caption=false,font=footnotesize,labelfont=rm,textfont=rm]{subfig}
\usepackage{textcomp}
\usepackage{stfloats}
\usepackage{url}
\usepackage{verbatim}
\usepackage{graphicx}
\usepackage{cite}
\usepackage{amssymb}

\usepackage{cases}
\usepackage{amsthm}
\usepackage{booktabs}
\usepackage{multirow}
\usepackage[
colorlinks=true,
linkcolor=black,
citecolor=blue,
urlcolor=black,
]{hyperref}
\usepackage{enumerate}
\usepackage{pifont}

\theoremstyle{plain}
\newtheorem{theorem}{Theorem}

\begin{document}

\title{Combining Adam and its Inverse Counterpart to Enhance Generalization of Deep Learning Optimizers}

\author{Tao Shi,
        Liangming Chen,
        Long Jin,
        and Mengchu Zhou
\thanks{This work was supported by the National Natural Science Foundation of China under Grants 62476115 and 62506148. (Tao Shi and Liangming Chen are co-first authors.) (Corresponding author: Long Jin.)}
\thanks{Tao Shi, Liangming Chen, and Long Jin are with the School of Information Science and Engineering, Lanzhou University, Lanzhou 730000, China (e-mails: taoshi1999@foxmail.com, lmchen@foxmail.com, jinlongsysu@foxmail.com).}
\thanks{Mengchu Zhou is with the Helen and John C. Hartmann Department of Electrical and Computer Engineering, New Jersey Institute of Technology, Newark, NJ 07102, USA (e-mail: zhou@njit.edu).}
}



\maketitle

\begin{abstract}
In the training of neural networks, adaptive moment estimation (Adam) typically converges fast but exhibits suboptimal generalization performance. A widely accepted explanation for its defect in generalization is that it often tends to converge to sharp minima. To enhance its ability to find flat minima, we propose its new variant named inverse Adam (InvAdam). The key improvement of InvAdam lies in its parameter update mechanism, which is opposite to that of Adam. Specifically, it computes element-wise multiplication of the first-order and second-order moments, while Adam computes the element-wise division of these two moments. This modification aims to increase the step size of the parameter update when the elements in the second-order moments are large and vice versa, which helps the parameter escape sharp minima and stay at flat ones. However, InvAdam's update mechanism may face challenges in convergence. To address this challenge, we propose dual Adam (DualAdam), which integrates the update mechanisms of both Adam and InvAdam, ensuring convergence while enhancing generalization performance. Additionally, we introduce the diffusion theory to mathematically demonstrate InvAdam's ability to escape sharp minima. Extensive experiments are conducted on image classification tasks and large language model (LLM) fine-tuning. The results validate that DualAdam outperforms Adam and its state-of-the-art variants in terms of generalization performance. The code is publicly available at https://github.com/LongJin-lab/DualAdam.
\end{abstract}


\section{Introduction}
\IEEEPARstart{O}{ptimizers} play an important role in training deep neural networks \cite{xu2025optimization, wen2024evolution}, significantly influencing their generalization performance \cite{zeng2008design, chen2024enhancing, jin2025ape}. Empirical and theoretical studies suggest a positive correlation between the generalization of neural networks and the flatness around the minimum in a loss landscape \cite{yang2023stochastic, zhou2020towards, zeng2025sharpness}. 
\begin{figure}[!h]
	\centering
	\includegraphics[width=1.0\linewidth]{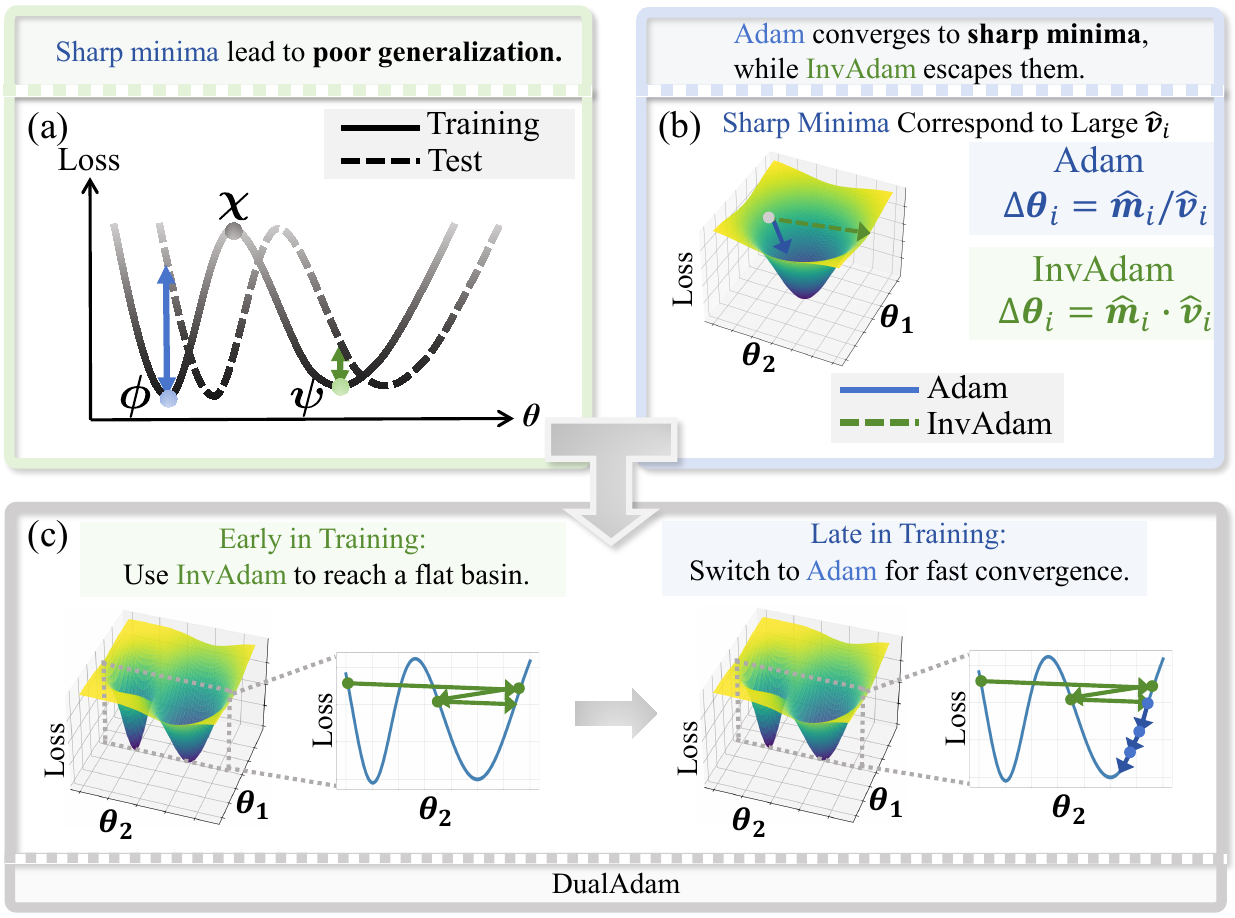}
	\caption{Mian idea of DualAdam. (a) The relationship between sharp minimum $\phi$ and flat one $\psi$. $\chi$ is the saddle point between $\phi$ and $\psi$; (b) The update mechanisms of Adam and InvAdam. $\boldsymbol{\hat{m}}$ and $\boldsymbol{\hat{v}}$ are the bias-corrected first- and second-order moments, respectively. Subscript $i$ represents the $i$-th element in the vector. $\Delta \boldsymbol{\theta}_i$ is the $i$-th element in parameter update in one iteration; and (c) The update mechanism of DualAdam.}
	\label{fig_first}
\end{figure}
As shown in Fig. \ref{fig_first}(a), a flat minimum in a loss landscape is a region where small perturbations on model parameter $\boldsymbol{\theta}$ lead to insignificant changes in the loss value. These regions are believed to be associated with better generalization performance because models that converge to flat minima are less sensitive to variations in the data \cite{xie2020diffusion}. Conversely, a sharp minimum is characterized by steep loss contours, making models more susceptible to overfitting and less generalizable to unseen data. Among the various optimizers, adaptive moment estimation (Adam) \cite{kingma2014adam} enjoys widespread popularity due to its fast convergence across different architectures \cite{wang2024provable, jia2024weight, zheng2025deep}. However, despite its rapid convergence, it often faces challenges in terms of generalization. A widely accepted explanation for its defect in generalization is that it often converges to sharp minima \cite{zhou2020towards, peng2024robust}. This is because the intuition behind its adaptive learning rate is to decrease the step size of the parameter update when the corresponding elements in the second-order moments are large and vice versa. This strategy helps alleviate oscillations in the parameter update and maintains effective update even when the gradient is close to zero \cite{jia2024weight}. While this approach enables Adam to converge quickly, it increases the probability of trapping into sharp minima \cite{zhou2020towards}. This occurs because the elements in the second-order moments around sharp minima are usually large, leading it to make small steps to navigate these regions. To address this limitation, we propose a new variant of Adam called inverse Adam (InvAdam). As shown in Fig. \ref{fig_first}(b), InvAdam's key idea lies in its update mechanism, where the first-order moment is multiplied element-wise by the second-order moment, instead of being divided as in Adam. This adjustment aims to increase the step size of the parameter update when the corresponding elements in the second-order moments are large and decrease it when the elements are small, which enhances InvAdam's ability to escape sharp minima and settle in flat ones. To mathematically demonstrate that it has a good ability to escape sharp minima, this work introduces the diffusion theory \cite{xie2020diffusion}. However, while the modification in InvAdam improves the ability to find flat minima, it poses challenges in terms of convergence. This occurs because an increase in the step size of the parameters update may cause the parameters to oscillate, making it difficult for the parameters to converge. To ensure the final convergence, we propose an optimizer named dual Adam (DualAdam), which integrates the update mechanisms of both Adam and InvAdam. As mentioned above, Adam is characterized by rapid convergence, while InvAdam is characterized by a good exploration capability for flat minima. As shown in Fig. \ref{fig_first}(c), by integrating these two distinct update mechanisms, DualAdam dynamically combines the strengths of both mechanisms to balance between convergence performance and generalization one. Specifically, DualAdam starts with InvAdam's update mechanism to explore flat minima and linearly transitions to Adam's update mechanism to ensure convergence. This transition is controlled by a switching rate, which determines the speed of the transition from InvAdam to Adam. In summary, this work aims to make the following novel contributions to the field of optimizers for deep learning:
\begin{enumerate}
\item We propose InvAdam, an optimizer with enhanced capabilities in escaping sharp minima;

\item We provide a theoretical foundation for InvAdam using the diffusion theory to analyze its ability to escape sharp minima; and

\item To overcome the convergence challenges of InvAdam, we propose DualAdam, which combines the update mechanisms of Adam and InvAdam to enhance Adam’s generalization performance while ensuring convergence.
\end{enumerate}

\section{Related work}

Adam is one of the most popular optimizers for training neural networks due to its effectiveness in achieving fast and stable convergence \cite{xiao2024adam, luo2024pseudo}. To enhance its generalization performance, its variants have been presented. Moreover, some studies explore the combination of multiple optimizers' update mechanisms to leverage their respective strengths, which provides new insights into optimization strategies.

\subsection{Adam Variants}

Adam with decoupled weight decay (AdamW) \cite{loshchilov2017decoupled} introduces decoupled weight decay (WD) regularization, which separates WD from the gradient. This decoupling addresses the issue of WD interacting with adaptive learning rates, leading to better generalization performance than Adam. Due to its convenience and effectiveness, decoupled WD regularization is widely adopted in adaptive methods to enhance generalization performance. However, this modification is relatively minor, and its effect on generalization performance is limited. Rectified Adam (RAdam) \cite{liu2019variance} attempts to rectify the variance of adaptive learning rates by incorporating a variance rectification term. This modification helps stabilize the learning rate during the early stages of training, when the number of iterations is insufficient to estimate the variance accurately. As a result, RAdam generalizes better than Adam in practice, reducing the probability of converging to poor local minima due to the initial variance of the learning rate. However, its rectification term may slow down the convergence. Nesterov-accelerated Adam (NAdam) \cite{dozat2016incorporating} incorporates Nesterov momentum into the Adam framework, enabling it to anticipate future gradient directions. This anticipation enhances the optimizer's ability to adjust updates effectively, guiding the parameter to escape sharp minima. However, this mechanism does not effectively improve generalization. Adaptive Nesterov momentum (Adan) \cite{xie2024adan} utilizes a Nesterov momentum estimation method to estimate stable and accurate first- and second-order moments of the gradient in adaptive gradient algorithms for acceleration. Experiments show that Adan outperforms Adam and its variants in terms of generalization performance. However, its update mechanism is complex, which may lead to implementation and tuning challenges. 

Unlike the optimizers mentioned above, DualAdam combines the update mechanisms of both InvAdam and Adam to improve generalization performance while ensuring convergence, dynamically balancing the strengths of both mechanisms through a linear switching mechanism.

\subsection{Switching between Two Update Mechanisms}

Some studies focus on combining different optimizers. Switching from Adam to stochastic gradient descent (SWATS) \cite{keskar2017improving} starts with Adam for fast convergence and switches to stochastic gradient descent (SGD) for better generalization. However, the challenge lies in determining the optimal switching point, which requires extra computation overhead and makes SWATS less efficient. Moreover, its effectiveness is debatable. Our experiments indicate that SWATS often exhibits poorer generalization than Adam, an observation consistent with the results in \cite{jin2025nong}. Unlike SWATS, DualAdam uses a linear switch method for a smooth transition from InvAdam to Adam. Jiang \textit{et al.} \cite{jiang2023adaptive} introduce an adaptive policy for employing sharpness-aware minimization (SAM) that combines SAM and SGD. However, its complexity poses challenges in implementation, tuning, and computation. The switching mechanism of DualAdam is straightforward to implement, since it requires only a switching rate to control the speed of transition from InvAdam to Adam. This simplicity makes DualAdam practical for large-scale optimization in deep learning. Multiple integral Adam (MIAdam) \cite{jin2025nong} aims to improve generalization by using a multiple integral moment mechanism in the early stages of training and switching to Adam after a specific number of epochs. This hard switching mechanism may lead to instability during the transition. Differently, DualAdam employs a linear switching mechanism that ensures a smooth transition from InvAdam to Adam, enhancing the performance without compromising stability throughout the training process. Experiments show that DualAdam also outperforms MIAdam in terms of generalization performance. Moreover, to the best of our knowledge, this work is the first to propose a linear switching mechanism between two distinct update rules for deep learning optimizers, which is enlightening and may be applied to other optimizers as well.

\section{Proposed Methods}

In this section, we first introduce the detailed implementation of DualAdam. Next, we analyze its computational complexity. Following this, we theoretically analyze the ability of Adam and InvAdam to escape sharp minima. Finally, we analyze the convergence of DualAdam.

\subsection{DualAdam}

The training procedure of a neural network can be characterized as an optimization problem \cite{chen2025multiscale}, i.e.,
\begin{align}
	\min_{\boldsymbol{\theta}} \mathcal{L}(\boldsymbol{x}, \boldsymbol{y}; \boldsymbol{\theta}),
	\label{eq_opt_prob}
\end{align}
where $\mathcal{L}(\boldsymbol{x}, \boldsymbol{y}; \boldsymbol{\theta})$ is a loss function; $\boldsymbol{\theta}$ is the parameter of a neural network; $\boldsymbol{x}$ and $\boldsymbol{y}$ are the input data and corresponding label. To solve \eqref{eq_opt_prob}, the gradient is calculated as: 
\begin{align}
	\boldsymbol{g}_{t,i} = \frac{\partial L(\boldsymbol{\theta}_{t,i})}{\partial \boldsymbol{\theta}_{t,i}},
	\label{eq_grad}
\end{align}
where $L(\boldsymbol{\theta}) = 1 / b \sum_{k=1}^{b} \mathcal{L}(\boldsymbol{x}_k, \boldsymbol{y}_k; \boldsymbol{\theta})$ is the loss function over one batch; subscript $k$ represents the $k$-th sample in a batch of data; $b$ is batch size; subscripts $t$ and $i$ denote the $t$-th iteration and the $i$-th element of the corresponding vector, respectively. Based on \eqref{eq_grad}, the first- and second-order moments can be expressed as:
\begin{subnumcases} { }
	\boldsymbol{m}_{t,i} = \beta_1 \boldsymbol{m}_{t-1,i} + (1 - \beta_1) \boldsymbol{g}_{t,i}, \label{eq_fir_moment}\\
	\boldsymbol{v}_{t,i} = \beta_2 \boldsymbol{v}_{t-1,i} + (1 - \beta_2) \boldsymbol{g}_{t,i}^2, \label{eq_sec_moment}
\end{subnumcases}
where $\boldsymbol{m}$ is the first-order moment; $\boldsymbol{v}$ is the second-order moment; $\beta_1$ and $\beta_2$ are the exponential decay rates used to adjust the first- and second-order moments, respectively. Furthermore, Adam's parameter update in one iteration can be calculated as:
\begin{align}
	\boldsymbol{u}_{t,i} = \frac{\boldsymbol{\hat{m}}_{t,i}}{\sqrt{\boldsymbol{\hat{v}}_{t,i}} + \epsilon}, 
	\label{eq_upd_adam}
\end{align}
where $\boldsymbol{u}_{\mathrm{Adam},t,i}$ is the $i$-th element of Adam's parameter update in the $t$-th iteration; $\boldsymbol{\hat{m}}_{t,i} = \boldsymbol{m}_{t,i} / (1 - \beta_1^t)$ and $\boldsymbol{\hat{v}}_{t,i} = \boldsymbol{v}_{t,i} / (1 - \beta_1^t)$ are the bias-corrected first-order and second-order moments, respectively; $\epsilon$ is a small positive value introduced to avoid division by zero.

In contrast to Adam, which divides $\boldsymbol{\hat{m}}_{t,i}$ by ${\sqrt{\boldsymbol{\hat{v}}_{t,i}}}$, InvAdam multiplies these two terms to compute the parameter update. This update mechanism is given by:
\begin{align}
\tilde{\boldsymbol{u}}_{t,i} = \boldsymbol{\hat{m}}_{t,i} \sqrt{\boldsymbol{\hat{v}}_{t,i}}, 
\label{eq_upd_InvAdam}
\end{align}
where $\tilde{\boldsymbol{u}}_{t,i}$ is the $i$-th element of InvAdam's parameter update in the $t$-th iteration. By comparing the two mechanisms in \eqref{eq_upd_adam} and \eqref{eq_upd_InvAdam}, it is evident that as $\boldsymbol{\hat{v}}_{t,i}$ increases, $|\boldsymbol{u}_{t,i}|$ decreases while $|\tilde{\boldsymbol{u}}_{t,i}|$ increases. This indicates that Adam slows down the parameter update when the corresponding elements in the second-order moments are large, whereas InvAdam speeds it up in such situations. This difference highlights the distinct goals of the adaptive learning rate mechanism in Adam and InvAdam: Adam's adaptive learning rate aims to alleviate oscillations in parameter update and maintains effective update even when the gradient is close to zero, while InvAdam's adaptive learning rate seeks to escape sharp minima and find flat ones.

By combining the update mechanisms of Adam and InvAdam described in \eqref{eq_upd_adam} and \eqref{eq_upd_InvAdam}, the parameter update of DualAdam in one iteration can be expressed as:
\begin{equation}
\bar{\bar{\boldsymbol{u}}}_{t,i} =  \alpha \tilde{\boldsymbol{u}}_{t,i} + (1 - \alpha) \boldsymbol{u}_{t,i},
\end{equation}
where $\bar{\bar{\boldsymbol{u}}}_{t,i}$ is the $i$-th element of DualAdam's parameter update in the $t$-th iteration; $\xi$ is the switching rate that controls the speed of transition from InvAdam to Adam; and $\alpha = \max(0, 1 - \xi t)$ represents the proportion of InvAdam's parameter update in DualAdam's one. As seen in Algorithm \ref{DualAdam}, the proportion of InvAdam's parameter update linearly decays as the number of iterations increases, while the proportion of Adam's parameter update increases. After a specific number of iterations, the proportion of the former decays to zero, meaning that DualAdam completely transitions to Adam. The ability to escape sharp minima effectively enables DualAdam to reach a relatively flat region of the loss landscape at the initial stage of training. The transition to Adam leverages Adam's strong convergence properties, ensuring the final convergence of DualAdam.

\begin{algorithm}[!ht]
\caption{DualAdam} \label{DualAdam}
\textbf{Given:} Learning rate: $\eta$;\\
exponential decay rates: $\beta_1$, $\beta_2$;\\
a small positive value: $\epsilon$;\\
switching rate: $\xi$;\\
\textbf{Initialize:} Step time $t \leftarrow$ 0;\\
the first-order moment $\boldsymbol{m}_{0,i}\leftarrow$ 0;\\
the second-order moment $\boldsymbol{v}_{0,i}\leftarrow$ 0;\\
\begin{algorithmic}[1]
\WHILE {stopping criterion is not met}
\STATE $t \leftarrow t+1$
\STATE Using \eqref{eq_grad} to get the gradient $\boldsymbol{g}_{t,i}$
\STATE Using \eqref{eq_fir_moment} to get the first-order moment $\boldsymbol{m}_{t,i}$
\STATE Using \eqref{eq_sec_moment} to get the second-order moment $\boldsymbol{v}_{t,i}$
\STATE $\boldsymbol{\hat{m}}_{t,i} \leftarrow \boldsymbol{m}_{t,i} / (1 - \beta_1^t)$
\STATE $\boldsymbol{\hat{v}}_{t,i} \leftarrow \boldsymbol{v}_{t,i} / (1 - \beta_2^t)$
\STATE $\alpha \leftarrow \mathrm{max}(0, 1 - t  \xi)$
\STATE $\boldsymbol{u}_{t,i} \leftarrow \boldsymbol{\hat{m}}_{t,i} / (\sqrt{\boldsymbol{\hat{v}}_{t,i}} + \epsilon) $
\STATE $\tilde{\boldsymbol{u}}_{t,i} \leftarrow \boldsymbol{\hat{m}}_{t,i} \sqrt{\boldsymbol{\hat{v}}_{t,i}}$
\STATE $\bar{\bar{\boldsymbol{u}}}_{t,i} \leftarrow  \alpha \tilde{\boldsymbol{u}}_{t,i} + (1 - \alpha) \boldsymbol{u}_{t,i}$
\STATE $\boldsymbol{\theta}_{t,i} \leftarrow \boldsymbol{\theta}_{t-1,i} - \eta \bar{\bar{\boldsymbol{u}}}_{t,i}$
\ENDWHILE
\end{algorithmic}
\end{algorithm}

\subsection{Computational Complexity Analysis of DualAdam} \label{complexity}

To quantitatively evaluate the computational efficiency of DualAdam, we decompose the floating-point operations (FLOPs) performed per parameter during a single training iteration. Let $p$ denote the number of model parameters. The computational process for DualAdam can be broken down into five distinct steps: \textbf{1)} Updating the first-order moment $m_{t,i}$ involves 2 multiplications and 1 addition, while the second-order moment $v_{t,i}$ requires 1 squaring, 2 multiplications, and 1 addition. This step contributes 7 FLOPs. \textbf{2)} Computing the bias-corrected estimates $\hat{m}_{t,i}$ and $\hat{v}_{t,i}$ requires 2 division operations (2 FLOPs). \textbf{3)} Calculating the standard term $u_{t,i}$ necessitates 1 square root, 1 addition (for $\epsilon$), and 1 division. Subsequently, the dual term $\tilde{u}_{t,i}$ reuses the square root term and requires 1 additional multiplication. This step contributes 4 FLOPs. \textbf{4)} The fusion of terms ($\bar{u}_t \leftarrow \alpha \tilde{u}_t + (1-\alpha)u_t$) involves 2 multiplications and 1 addition, contributing 3 FLOPs. \textbf{5)} The final weight update requires 1 multiplication by the learning rate and 1 subtraction, adding 2 FLOPs. Aggregating these steps, DualAdam incurs a total theoretical cost of $18p$ FLOPs per iteration. In comparison, Adam follows an identical procedure for steps \textbf{1)} and \textbf{2)} but simplifies steps \textbf{3)} through \textbf{5)} into a single update operation involving 1 square root, 1 addition, 1 division, 1 multiplication, and 1 subtraction, resulting in a total of $14p$ FLOPs. While DualAdam introduces a marginal overhead of roughly $4p$ FLOPs, this increase is negligible in the context of large-scale deep learning. According to standard scaling laws \cite{kaplan2020scaling}, the dominant computational load stems from the forward and backward propagation, which scale as approximately $6bp$ FLOPs (where $b$ is the batch size). Consequently, the ratio of the optimizer's additional overhead to the total training compute is proportional to $O(1/b)$. For typical batch sizes ($b \gg 1$), the extra computational overhead introduced by DualAdam constitutes a vanishingly small fraction of the total computational budget.

Moreover, the extra computation is only necessary during the early stages of DualAdam. As $\alpha$ decays to zero, DualAdam transitions fully into Adam, with the only remaining extra computation being the calculation of $\alpha$ and checking if it has reached zero. In our experiments on CIFAR-10 and CIFAR-100, we set the switching rate $\xi$ to $8 \times 10^{-5}$, resulting in a complete transition after 12,500 iterations. Given a batch size of 128 and a total of 200 epochs, the total number of iterations will be 78,200. Thus, the majority of the extra computation is required for only a brief portion (i.e., 15.98\%) of the training process. Since the computational efficiency of Adam has been thoroughly studied by \cite{kingma2014adam}, we do not delve into further theoretical analysis here. However, we provide experimental comparisons of the computational time for DualAdam and Adam, demonstrating that their computational time is not much different. For example, when training ResNet-18 on CIFAR-100 with a batch size of 128 for 200 epochs, Adam takes 2197 seconds, while DualAdam takes 2214 seconds, indicating a negligible difference in computational time.

\subsection{Theoretical Analysis of Ability to Escape Sharp Minima} \label{theoretical_analysis}

In this section, we present a mathematical analysis of InvAdam's ability to escape sharp minima using the diffusion theory framework and compare this ability to that of Adam. The diffusion theory employs the mean escape time as a metric to quantify the ability of optimizers to escape sharp minima. Specifically, a small mean escape time corresponds to a strong ability to escape sharp minima. For the analysis of the mean escape time, three assumptions are stated by following \cite{xie2020diffusion}:
\newline \newline
\textbf{Assumption 1 (Second Order Taylor Approximation).} \textit{The loss function around point $\boldsymbol{\theta}_0$ can be approximately written as}
{
\small
\begin{align}
\nonumber
L(\boldsymbol{\theta}) = L(\boldsymbol{\theta}_0)+\boldsymbol{g}(\boldsymbol{\theta}_0)(\boldsymbol{\theta} - \boldsymbol{\theta}_0) + \frac{1}{2} (\boldsymbol{\theta} - \boldsymbol{\theta}_0)^\top H(\boldsymbol{\theta}_0) (\boldsymbol{\theta} - \boldsymbol{\theta}_0).
\end{align}
}\textit{where $\boldsymbol{g}(\boldsymbol{\theta}_0)$ is the gradient at $\boldsymbol{\theta}_0$ and $H(\boldsymbol{\theta}_0)$ is the Hessian matrix at $\boldsymbol{\theta}_0$.}
\newline \newline
\textbf{Assumption 2 (Quasi-Equilibrium Approximation).} \textit{A system is in quasi-equilibrium near minima, which means the distribution of the parameter $\boldsymbol{\theta}$ can be described as a Boltzmann distribution.}
\newline \newline
\textbf{Assumption 3 (Low-Temperature Approximation).} \textit{The escape dynamics is primarily governed by the shape and height of the potential barrier rather than thermal noise, i.e., the gradient noise is small in our context.}
\newline

If Assumption 1--3 hold, as illustrated in Fig. \ref{fig_first}(a), we can formulate Adam's mean escape time from sharp minimum $\phi$ through the saddle point $\chi$ as
\begin{align}
\mathrm{log}(\tau) = \mathrm{O} \left( \frac{2 \sqrt{b} \Delta L}{\eta \sqrt{H_{\boldsymbol{\phi e}}}} \right),
\label{eq_log_adam_escape}
\end{align}
where $\tau$ is the mean escape time of Adam, $H_{\boldsymbol{\phi e}}$ is the eigenvalue of the Hessian matrix at the minimum $\boldsymbol{\phi}$ corresponding to the escape direction $\boldsymbol{e}$, and $\Delta L = L(\chi)-L(\phi)$. The detailed proof of \eqref{eq_log_adam_escape} can be found in \cite{xie2022adaptive}. Furthermore, for the mean escape time of InvAdam, we can derive Theorem \ref{th_inverse_escape} as follows:
\begin{theorem} \label{th_inverse_escape}
 If Assumptions 1--3 hold, and the dynamics is governed by InvAdam, then the mean escape time from minimum $\boldsymbol{\phi}$ to the outside of $\boldsymbol{\phi}$ is
{
\small
\begin{align}
\nonumber
\tilde{\tau} &= \pi \left( \sqrt{1 + \frac{4 \eta b \lvert H_{\boldsymbol{\phi e}} \rvert ^ \frac{3}{2}}{\left( 1 - \beta_1 \right) \sqrt{b}}} + 1 \right)
\frac{\lvert \mathrm{det}(H_{\boldsymbol{\phi}}^{-1} H_{\boldsymbol{\chi}}) \rvert ^ {-\frac{1}{4}}}{\lvert H_{\boldsymbol{\chi e}} \rvert}  \\ 
&\exp \left( \frac{2 b ^ \frac{3}{2} \Delta L}{\eta} \left( \frac{s}{H_{\boldsymbol{\phi e}} ^ \frac{3}{2}} + \frac{1-s}{\lvert H_{\boldsymbol{\chi e}} \rvert ^ \frac{3}{2}} \right) \right),
\label{eq_inverse_escape}
\end{align}
}where $H_{\boldsymbol{\chi e}}$ is the eigenvalue of the Hessian matrix of the loss function at saddle point $\boldsymbol{\chi}$ along escape direction $\boldsymbol{e}$ and $s \in (0,1)$ is a path-dependent coefficient.
\end{theorem}
$\mathbf{Proof.}$ The diffusion theory views the process of escaping sharp minima as a Kramers escape problem \cite{xie2020diffusion}. The problem is a classic one in statistical mechanics, describing how a particle escapes from a potential well under the influence of thermal noise \cite{Berera2019Formulating}. As is shown in Fig. \ref{fig_first}(a), the problem is introduced here to describe a process of parameter $\boldsymbol{\theta}$ escaping from sharp minimum $\boldsymbol{\phi}$, crossing saddle point $\boldsymbol{\chi}$, and finally converging to flat minimum $\boldsymbol{\psi}$. The mean escape time is used in statistical physics and stochastic processes to measure the ability of particles to escape from a potential well in the Kramers escape problem \cite{xie2020diffusion}. The potential well can be analogized as a minimum of the loss landscape during the model training, and the position of the particle can be analogized as the parameter $\boldsymbol{\theta}$.
The Langevin equation \cite{pacheco2024langevin} is a stochastic differential equation used to describe the dynamics of a particle under the influence of both deterministic forces and random thermal noise. It is the theoretical basis of the Kramers escape problem and can be written as: 
\begin{align}
\gamma \frac{\mathrm{d}\boldsymbol{\theta}}{\mathrm{d}t} = - \frac{\mathrm{d}U(\boldsymbol{\theta})}{\mathrm{d}\boldsymbol{\theta}} + r(t),
\label{eq_langevin}
\end{align}
where $\gamma$ is a damping coefficient, $U(\boldsymbol{\theta})$ is the potential energy function, and $r(t) \sim \mathcal{N}(0, \sigma^2 I)$ ($\sigma$ is a constant denoting standard deviation and $I$ is an identity matrix) is a random force representing thermal noise. The diffusion theory analogizes $U(\boldsymbol{\theta})$ to $L(\boldsymbol{\theta})$, and $r(t)$ to the gradient noise, allowing the use of the Kramers escape problem to study the process of parameter $\boldsymbol{\theta}$ escaping from the minima. Similar to  \eqref{eq_langevin}, the dynamics of SGD can be written as:
\begin{align}
\mathrm{d}\boldsymbol{\theta} = - \nabla L(\boldsymbol{\theta})\mathrm{d}t + (\eta C(\boldsymbol{\theta}))^\frac{1}{2} \mathrm{d}W_t,
\label{eq_sgd_langevin}
\end{align}
where $\mathrm{d}t$ is the differential of time, $\mathrm{d}W_t \sim \mathcal{N}(0, I \mathrm{d}t)$, and $C(\boldsymbol{\theta})$ is a gradient noise covariance matrix. The gradient noise is the difference between the stochastic gradient over one batch and the true gradient over the entire training dataset. It can be characterized by the gradient noise covariance matrix $C(\boldsymbol{\theta})$. $C(\boldsymbol{\theta})$ near a critical point can be calculated as
{
\small
\begin{align}
\nonumber
C(\boldsymbol{\theta}) &= \frac{1}{b} \left( \frac{1}{m} \sum_{j=1}^{m} \nabla L(\boldsymbol{\theta}, \boldsymbol{x}_j) \nabla L(\boldsymbol{\theta}, \boldsymbol{x}_j) ^ \top - \nabla L(\boldsymbol{\theta}) \nabla L(\boldsymbol{\theta}) ^ \top \right) \\
&\approx \frac{1}{bm} \sum_{j=1}^{m} \nabla L(\boldsymbol{\theta}, \boldsymbol{x}_j) \nabla L(\boldsymbol{\theta}, \boldsymbol{x}_j) ^ \top,
\label{eq_gradient_noise}
\end{align}
}where $m$ is the total number of data samples in the dataset, $\boldsymbol{x}_j$ is a data sample from the dataset, and $b$ is batch size. Based on \eqref{eq_gradient_noise}, \cite{xie2020diffusion} further validates both theoretically and empirically that $C(\boldsymbol{\theta}) \approx H(\boldsymbol{\theta}) / b $ during the whole training process. 

The Fokker-Planck equation \cite{boffi2023probability} is a partial differential equation that describes the time evolution of the probability density function of a particle's velocity under the influence of forces. It is widely used to model the dynamics of systems affected by random fluctuations. The escape rate can be derived by solving the Fokker-Planck equation for a particle in a double-well potential, where one well represents the trapped state and the other represents the escaped state. The solution helps determine how the probability density of the particle in a certain state evolves over time and how likely it is to cross the barrier separating these states. Therefore, to calculate the mean escape time of parameter $\boldsymbol{\theta}$ in the Kramers escape problem, the Fokker-Planck equation can be written as
\begin{align}
\frac{\partial P(\boldsymbol{\theta}, t)}{\partial t} = \nabla \cdot (P(\boldsymbol{\theta}, t) \nabla L(\boldsymbol{\theta})) + \nabla \cdot \nabla D(\boldsymbol{\theta}) P(\boldsymbol{\theta}, t),
\label{eq_fokker}
\end{align}
where $P(\boldsymbol{\theta}, t)$ is the probability density function of  parameter $\boldsymbol{\theta}$ over time, $\nabla \cdot$ is a divergence operator, $D(\boldsymbol{\theta}) = \eta C(\boldsymbol{\theta}) / 2$ is the diffusion matrix and  $C(\boldsymbol{\theta})$ is the gradient noise covariance matrix. 

Gauss's divergence theorem states that the surface integral of a vector field over a closed surface (the flux) is equal to the volume integral of the divergence over the enclosed region. We denote the mean escape time as $\hat{\tau}$, the escape rate as $\omega$, and the probability current as $J$.  Applying Gauss's divergence theorem to the Fokker-Planck equation, we obtain:
\begin{align}
\frac{\partial P(\boldsymbol{\theta}, t)}{\partial t} = - \nabla \cdot J(\boldsymbol{\theta}, t).
\label{eq_J}
\end{align}
Then, the mean escape time can be formulated as
\begin{align}
\hat{\tau} = \frac{1}{\omega} = \frac{P(\boldsymbol{\theta} \in V_{\boldsymbol{\phi}})}{\int_{S_{\boldsymbol{\phi}}} J \mathrm{d}S},
\label{eq_tau}
\end{align}
where $P(\boldsymbol{\theta} \in V_{\boldsymbol{\phi}}) = \int_{V_{\boldsymbol{\phi}}} P(\boldsymbol{\theta}) \mathrm{d}V$ represents the probability of parameter $\boldsymbol{\theta}$  within volume $V_{\boldsymbol{\phi}}$ enclosing minimum $\boldsymbol{\phi}$, $J$ denotes the probability current arising from probability source $P(\boldsymbol{\theta} \in V_{\boldsymbol{\phi}})$, and $j = \int_{S_{\boldsymbol{\phi}}} J \mathrm{d}S$ represents the probability flux, calculated as the surface integral of the probability current over surface $S_{\boldsymbol{\phi}}$. $S_{\boldsymbol{\phi}}$ surrounds minimum $\boldsymbol{\phi}$, and $V_{\boldsymbol{\phi}}$ is the volume enclosed by $S_{\boldsymbol{\phi}}$.
	
For the analysis of the momentum mechanism's dynamics, we first introduce the heavy ball method to describe the continuous-time momentum dynamics, i.e.,
\begin{align}
	\begin{cases}
	\boldsymbol{m}_t = \kappa_1 \boldsymbol{m}_{t-1} + \kappa_2 \boldsymbol{g}_t, \\
	\boldsymbol{\theta}_{t+1} = \boldsymbol{\theta}_t - \eta \boldsymbol{m}_t,
	\end{cases}
	\label{eq_momen}
	\end{align}
	where $\kappa_1$ and $\kappa_2$ are the hyperparameters. If $\kappa_2 = 1$, it is the SGD-style momentum, and if $\kappa_2 = 1 - \kappa_1$, it is the Adam-style momentum. Then, the motion equation with the damping coefficient $\gamma$ and the mass $M$ can be written as
	\begin{align}
	\begin{cases}
	\boldsymbol{r}_t = (1 - \gamma \eta) \boldsymbol{r}_{t-1} + \frac{\boldsymbol{F}}{M} \eta, \\
	\boldsymbol{\theta}_{t+1} = \boldsymbol{\theta}_t + \eta \boldsymbol{r}_t, 
	\end{cases}
	\label{eq_motion}
	\end{align}
	where $\boldsymbol{r}_t = - \boldsymbol{m}_t$, $\boldsymbol{F} = \boldsymbol{g}_t$, $1 - \gamma \eta = \kappa_1 = \beta_1$, and $\eta/M = \kappa_2 = 1 - \beta_1$. Then, we can write the differential form of the motion equation, which describes the dynamics of the momentum as
	\begin{align}
	M \ddot{\boldsymbol{\theta}} = - \gamma M \dot{\boldsymbol{\theta}} + \boldsymbol{F},
	\label{eq_de_motion}
	\end{align}
	where $\ddot{\boldsymbol{\theta}} = \mathrm{d}^2 \boldsymbol{\theta} / \mathrm{d}t^2$ and $\dot{\boldsymbol{\theta}} = \mathrm{d}\boldsymbol{\theta} / \mathrm{d}t$. As $\boldsymbol{F}$ corresponds to the stochastic gradient term, we can rewrite \eqref{eq_de_motion} as 
	\begin{align}
	M \ddot{\boldsymbol{\theta}} = - \gamma M \dot{\boldsymbol{\theta}} - \frac{\partial L(\boldsymbol{\theta})}{\partial \boldsymbol{\theta}} dt + (2D(\boldsymbol{\theta}))^\frac{1}{2}dW_t.
\end{align}
The Fokker-Planck equation in the phase space ($\boldsymbol{\theta}-\boldsymbol{\dot{\theta}}$ space) is given as
\begin{equation}
\begin{aligned}
&\frac{\partial P(\boldsymbol{\theta}, \boldsymbol{\dot{\theta}}, t)}{\partial t} \\
&= - \nabla_{\boldsymbol{\theta}} \cdot (\boldsymbol{\dot{\theta}} P(\boldsymbol{\theta}, \boldsymbol{\dot{\theta}}, t) ) \\
&+ \nabla_{\boldsymbol{\dot{\theta}}} \cdot (\omega \boldsymbol{\dot{\theta}} + M^{-1} \nabla_{\boldsymbol{\theta}} L(\nabla_{\boldsymbol{\theta}}) ) P(\boldsymbol{\theta}, \boldsymbol{\dot{\theta}}, t) \\
&+ \nabla_{\boldsymbol{\dot{\theta}}} M^{-2} D(\boldsymbol{\theta}) \cdot \nabla_{\boldsymbol{\dot{\theta}}} P(\boldsymbol{\theta}, \boldsymbol{\dot{\theta}}, t).
\end{aligned}	
\label{eq_fokker_momen}
\end{equation}
Under Assumption 2 (Quasi-Equilibrium Approximation), the distribution around minimum $\boldsymbol{\phi}$ is 
\begin{align}
P(\boldsymbol{\theta}) = P(\boldsymbol{\phi}) \mathrm{exp}(-\frac{\gamma M}{2} (\boldsymbol{\theta} - \boldsymbol{\phi})^\top (D_{\boldsymbol{\phi}}^{-\frac{1}{2}} H_{\boldsymbol{\phi}} D_{\boldsymbol{\phi}}^{-\frac{1}{2}}) (\boldsymbol{\theta} - \boldsymbol{\phi})),
\end{align}
where $D_{\boldsymbol{\phi}}$ is the diffusion matrix $D$ at the minimum $\boldsymbol{\phi}$, and $H_{\boldsymbol{\phi}}$ is the Hessian matrix at $\boldsymbol{\phi}$. Then we have
\begin{equation}
\begin{aligned}
&P(\boldsymbol{\theta} \in V_{\boldsymbol{\phi}}) \\
&= \int_{\boldsymbol{\theta} \in V_{\boldsymbol{\phi}}} P(\boldsymbol{\theta}) \mathrm{d}V \\
& = P(\boldsymbol{\phi}) \int_{\boldsymbol{\theta} \in V_{\boldsymbol{\phi}}} \\
&\mathrm{exp} \left( -\frac{\gamma M}{2} (\boldsymbol{\theta} - \boldsymbol{\phi})^\top (D_{\boldsymbol{\phi}}^{-\frac{1}{2}} H_{\boldsymbol{\phi}} D_{\boldsymbol{\phi}}^{-\frac{1}{2}}) (\boldsymbol{\theta} - \boldsymbol{\phi}) \right) \mathrm{d}V,\\
&\approx P(\boldsymbol{\phi}) \int_{\boldsymbol{\theta} \in (-\boldsymbol{\infty}, +\boldsymbol{\infty})} \\
&\mathrm{exp} \left( -\frac{\gamma M}{2} (\boldsymbol{\theta} - \boldsymbol{\phi})^\top (D_{\boldsymbol{\phi}}^{-\frac{1}{2}} H_{\boldsymbol{\phi}} D_{\boldsymbol{\phi}}^{-\frac{1}{2}}) (\boldsymbol{\theta} - \boldsymbol{\phi}) \right) \mathrm{d}V,\\
&= P(\boldsymbol{\phi}) \frac{(2 \pi \gamma M)^{\frac{n}{2}}}{\mathrm{det} (D_{\boldsymbol{\phi}}^{-1} H_{\boldsymbol{\phi}})^{\frac{1}{2}}}.
\end{aligned}
\label{eq_p_V_phi}
\end{equation}
For calculating probability flux $j = \int_{S_{\boldsymbol{\phi}}} J \mathrm{d}S$, we first reduce \eqref{eq_fokker_momen} to a space-dependent Smoluchowski-like equation, which is extended by an effective diffusion correction \cite{xie2022adaptive}:
\begin{align}
\hat{D}_i (\boldsymbol{\theta}) = D_i (\boldsymbol{\theta}) \left(1 - \sqrt{1-\frac{4 H_i(\boldsymbol{\theta})}{\gamma^2 M}} \right) \left( \frac{2 H_i(\theta)}{\gamma^2 M} \right)^{-1},
\label{eq_d_i}
\end{align}
where $D_i (\boldsymbol{\theta})$ is the $i$-th eigenvalue of the diffusion matrix $D(\boldsymbol{\theta})$ and $H_i(\boldsymbol{\theta})$ is the $i$-th eigenvalue of the Hessian matrix $H(\boldsymbol{\theta})$. Since our analysis is confined to the escape direction (an eigenvector direction), we simplify the problem by using the one-dimensional form of the Smoluchowski equation along this direction. For SGD, probability current $J_{1d}$ can be expressed as the Smoluchowski equation in position space:
\begin{align}
J_{1d} = D(\theta) \mathrm{exp} \left( \frac{-L(\theta)}{T} \right) \nabla \left( \mathrm{exp} \left( \frac{L(\theta)}{T} \right) P(\theta) \right),
\label{eq_sgd_J}
\end{align}
where $T$ is the temperature coefficient. Furthermore, for the momentum dynamics, probability current $\hat{J}_{1d}$ can be expressed by transforming \eqref{eq_sgd_J} into the position-space Smoluchowski-like form with the effective diffusion correction:
\begin{align}
\hat{J}_{1d} = \hat{D}(\theta) \mathrm{exp} \left( \frac{-L(\theta)}{\hat{T}} \right) \nabla \left( \mathrm{exp} \left( \frac{L(\theta)}{\hat{T}} \right) P(\theta) \right),
\label{eq_momentum_J}
\end{align}
where $\hat{T}=T/(\gamma M)$ and $\hat{D}(\theta)$ is the effective diffusion matrix with the eigenvalues corrected by \eqref{eq_d_i}. 

Let $c$ represent a point on escape direction $e$ from minimum $\phi$ to saddle point $\chi$, where $L(c) = (1-s)L(\phi)+sL(\chi)$. Temperature $T_{\phi}$ determines the path $\phi$ to $c$, and temperature $T_{\chi}$ determines the path $c$ to $\chi$. Then we have
\begin{equation}
\begin{aligned}
\nabla \left( \mathrm{exp} \left( \frac{L(\theta)-L(c)}{T} \right) P(\theta) \right) \\
= \hat{J}_{1d} \hat{D}^{-1} \mathrm{exp} \left( \frac{L(\theta)-L(s)}{T} \right).
\end{aligned}
\label{eq_form_J}
\end{equation}
For the left side of \eqref{eq_form_J}, we have
\begin{equation}
\begin{aligned}
&\nabla \left( \mathrm{exp} \left( \frac{L(\theta)-L(c)}{T} \right) P(\theta) \right) \\
&= \int_{\phi}^{\chi} \frac{\partial}{\partial \theta} \left( \mathrm{exp} \left( \frac{L(\theta) - L(c)}{T} \right) P(\theta) \right) \mathrm{d} \theta \\
&= \int_{\phi}^{c}  \frac{\partial}{\partial \theta} \left( \mathrm{exp} \left( \frac{L(\theta) - L(c)}{T_{\chi}} \right) P(\theta) \right) \mathrm{d} \theta \\
&+ \int_{c}^{\chi}  \frac{\partial}{\partial \theta} \left( \mathrm{exp} \left( \frac{L(\theta) - L(c)}{T_{\phi}} \right) P(\theta) \right) \mathrm{d} \theta \\
&= \left( P(c) - \mathrm{exp} \left( \frac{L(\chi) - L(c)}{T_\chi} \right) P(\chi) \right) \\
&+ (0 - P(c)) \\
&= -\mathrm{exp} \left( \frac{L(\chi) - L(c)}{T_\chi} \right) (\chi).
\end{aligned}
\end{equation}
For the right side of \eqref{eq_form_J}, we have
\begin{equation}
\begin{aligned}
	&\hat{J}_{1d} \hat{D}^{-1} \mathrm{exp} \left( \frac{L(\theta)-L(s)}{T} \right) \\
	&=-\hat{J}_{1d} \int_{\phi}^{\chi} \hat{D}^{-1}  \mathrm{exp} \left( \frac{L(\theta) - L(c)}{T} \right) \mathrm{d} \theta.
\end{aligned}
\end{equation}
Then we have 
\begin{align}
\hat{J}_{1d} = \frac{\mathrm{exp} \left( \frac{L(\chi) - L(c)}{T_\chi} \right) (\chi)}{\int_{\phi}^{\chi} \hat{D}^{-1}  \mathrm{exp} \left( \frac{L(\theta) - L(c)}{T} \right) \mathrm{d} \theta}.
\end{align}
Based on the formula of the one-dimensional probability current and flux, the high-dimensional flux escaping through $\boldsymbol{\chi}$ can be written as:
{
\small
\begin{equation}
\begin{aligned}
&\int_{S_{\boldsymbol{\chi}}} J \mathrm{d} S =\hat{J}_{1d} \!\! \int_{S_{\boldsymbol{\chi}}}\\
&\quad \quad \mathrm{exp} \!\! \left( \!\! -\frac{\gamma M}{2} (\boldsymbol{\theta} - \boldsymbol{\chi})^\top (D_{\boldsymbol{\chi}}^{-\frac{1}{2}} H_{\boldsymbol{\chi}} D_{\boldsymbol{\chi}}^{-\frac{1}{2}})^{\perp \boldsymbol{e}} (\boldsymbol{\theta} \! - \! \boldsymbol{\chi}) \! \right) \!\!\mathrm{d} S \\
&= \hat{J}_{1d} \frac{(2 \pi \gamma M)^{\frac{n-1}{2}}}{(\prod_{i \neq e} D_{\boldsymbol{\chi i}^{-1} H_{\boldsymbol{\chi i}}})^{\frac{1}{2}}}\\
&= \!\! \frac{\mathrm{exp} \left( \frac{L(\boldsymbol{\phi})-L(\boldsymbol{c})}{T_{\boldsymbol{\phi e}}} \right) P(\boldsymbol{\phi}) (2\pi\gamma M)^{\frac{n-1}{2}}} {\hat{D}_{\boldsymbol{\chi e}}^{-1} \mathrm{exp} \left( \frac{L(\boldsymbol{\chi})-L(\boldsymbol{c})}{T_{\boldsymbol{\chi e}}} \right) \sqrt{\frac{2\pi T_{\boldsymbol{\chi e}}}{|H_{\boldsymbol{\chi e}|}} } \left(\prod_{i \neq e} (D_{\boldsymbol{\chi i}}^{-1} H_{\boldsymbol{\chi i}})\right)^{\frac{1}{2}}}.
\end{aligned}	
\end{equation}
}where $(\cdot)^{\perp \boldsymbol{e}}$ represents the directions perpendicular to the escape direction $\boldsymbol{e}$. Then the mean escape time from minimum $\boldsymbol{\phi}$ to the outside of $\boldsymbol{\phi}$ is given as:
\begin{equation}
\begin{aligned}
\hat{\tau} &= \frac{P(\boldsymbol{\theta} \in V_{\boldsymbol{\phi}})}{\int_{S_{\boldsymbol{\chi}}} J \mathrm{d} S} \\
&= \pi \left( \sqrt{1 + \frac{4}{\gamma^2 M} \left| H_{\boldsymbol{\chi e}} \right|} + 1 \right) \frac{1}{\left| H_{\boldsymbol{\chi e}} \right|} \\
&\exp \left( \frac{2 \gamma M b \Delta L}{\eta} \left( \frac{s}{H_{\boldsymbol{\phi e}}} + \frac{1 - s}{\left| H_{\boldsymbol{\chi e}} \right|} \right) \right),
\label{eq_momen_escape}
\end{aligned}	
\end{equation}
where subscript $\boldsymbol{e}$ indicates the escape direction, $H_{\boldsymbol{\phi e}}$ and $H_{\boldsymbol{\chi e}}$ are the eigenvalues of the Hessian matrix of the loss function at minimum $\boldsymbol{\phi}$ and saddle point $\boldsymbol{\chi}$ along escape direction $\boldsymbol{e}$, and $\Delta L = L(\boldsymbol{\chi}) - L(\boldsymbol{\phi})$. To prove Theorem 1, we can replace standard learning rate $\eta$ with adaptive learning rate $\tilde{\eta}$ since InvAdam relies heavily on the momentum mechanism in the process of computing an update. The adaptive learning rate $\tilde{\eta}$ in the InvAdam can be written as
\begin{align}
\tilde{\eta} = \sqrt{V} \eta,
\label{eq_lr}
\end{align}
where $V = \mathbb{E} \left( \boldsymbol{g}_t \boldsymbol{g}_t^\top \right)$ is a matrix. This method of calculating $V$ is used by \cite{xie2022adaptive} to simplify the theoretical analysis.  $V = \mathbb{E} \left( \boldsymbol{g}_t \boldsymbol{g}_t^\top \right) = C \left( \boldsymbol{\theta} \right) = H / b$ approximately holds near critical points, where $H$ is the Hessian matrix. By replacing $\eta$ in \eqref{eq_momen_escape} with $\tilde{\eta}$ and setting $\gamma M = 1$ to simplify the derivation, we can get the InvAdam's mean escape time as:
\begin{equation}
\begin{aligned}
	&\tilde{\tau} = \pi \left( \sqrt{1 + \frac{4 \eta b \lvert H_{\boldsymbol{\chi e}} \rvert ^ \frac{3}{2}}{\left( 1 - \beta_1 \right) \sqrt{b}}} + 1 \right) \frac{\lvert \mathrm{det}(H_{\boldsymbol{\phi}}^{-1} H_{\boldsymbol{\chi}}) \rvert ^ {-\frac{1}{4}}}{\lvert H_{\boldsymbol{e}} \rvert} \\
	&\exp \left( \frac{2 b ^ \frac{3}{2} \Delta L}{\eta} \left( \frac{s}{H_{\boldsymbol{\phi e}} ^ \frac{3}{2}} + \frac{1-s}{\lvert H_{\boldsymbol{\chi e}} \rvert ^ \frac{3}{2}} \right) \right). 
\end{aligned}
\end{equation}
The proof is thus completed.\hfill $\blacksquare$

Based on Theorem \ref{th_inverse_escape}, we can write $\mathrm{log}\left( \tilde{\tau} \right)$ as

\begin{align}
\mathrm{log} (\tilde{\tau})  = \mathrm{O} \left( \frac{2 \sqrt{b} \Delta L}{\eta H_{\boldsymbol{\phi e}}^\frac{3}{2}} \right).
\label{eq_log_inverse_escape}
\end{align}

Based on \eqref{eq_log_adam_escape} and \eqref{eq_log_inverse_escape}, we can write the further simplified approximations as follows:
\begin{align}
\begin{cases}
\mathrm{log} (\tau) = \mathrm{O} (H_{\boldsymbol{\phi e}}^{-\frac{1}{2}}), \\
\mathrm{log} (\tilde{\tau}) = \mathrm{O} (H_{\boldsymbol{\phi e}}^{-\frac{3}{2}}).
\end{cases}
\label{eq_log_sim}
\end{align}
In the diffusion theory, the loss landscape's sharpness around minimum $\boldsymbol{\phi}$ is reflected by $H_{\boldsymbol{\phi e}}$, which is the eigenvalue of the Hessian matrix corresponding to the escape direction along an eigenvector. The relationship between the Hessian matrix’s eigenvalues and the sharpness of the loss landscape is also studied by \cite{dinh2017sharp} and \cite{bosman2023empirical}. Therefore, Eq. \eqref{eq_log_sim} shows that as $H_{\boldsymbol{\phi e}}$ increases, i.e., the escaping direction becomes sharper, the mean escape time of InvAdam decreases faster than Adam, indicating that InvAdam has a better ability to escape sharp minima compared to Adam. 

As mentioned above, DualAdam combines the update mechanism of InvAdam with that of Adam to strike a balance between convergence and generalization. As shown in Algorithm \ref{DualAdam}, $\alpha \in [0, 1]$ is the ratio of InvAdam, which decays linearly from 1 to 0 as the number of iterations increases during training. This means that at the initial stage of training, DualAdam's parameter update is entirely based on InvAdam's, which has a better ability to escape sharp minima than Adam. This allows InvAdam to explore the loss landscape more extensively in the early stages of the training, helping parameters reach regions near relatively flat minima. However, as the number of iterations increases, DualAdam's update transitions linearly to Adam's. This transition ensures that DualAdam can effectively converge in the later stages of the training, aligning with our goal of balancing generalization and convergence. 

\subsection{Convergence Analysis}
In analyzing convergence properties \cite{fan2025convergence}, foundational studies on neural network stability \cite{zeng2004stability, zeng2003global, gao2025stability} provide a concrete theoretical basis, as neural network stability supports and enhances convergence rates in optimization processes. Given that the convergence of Adam has been rigorously analyzed by \cite{wang2024provable} and \cite{li2024convergence}, and that DualAdam switches to Adam completely in later training stages, its convergence is guaranteed. Therefore, further proof is omitted here. However, it is important to note that InvAdam, when used solely, may encounter convergence challenges, to be demonstrated in the experiments. 

\section{Simulations and Experiments}

In this section, we perform extensive simulations and experiments to validate the performance of DualAdam and compare it with other optimizers, including Adam \cite{kingma2014adam}, AdamW \cite{loshchilov2017decoupled}, RAdam \cite{liu2019variance}, SWATS \cite{keskar2017improving}, NAdam \cite{dozat2016incorporating}, Adan \cite{xie2024adan}, and MIAdam \cite{jin2025nong}. Firstly, we conduct numerical simulations on 2-parameter loss landscapes \cite{yang2023towards} to evaluate InvAdam's ability to escape sharp minima. Secondly, image classification experiments are performed using ResNet-18, ResNet-50 \cite{he2016deep}, VGG-16 \cite{simonyan2014very}, and ViT-Small \cite{touvron2022deit} on CIFAR-10, CIFAR-100 \cite{Qiao2025Gradient}, Tiny ImageNet \cite{su2025generative}, and ImageNet-1k \cite{deng2009imagenet}. Next, the fast computation method of Hessian information of loss landscapes presented in \cite{yao2020pyhessian} is used to compare the density of the Hessian matrix's eigenvalues around the solutions obtained by Adam and DualAdam. Moreover, we present the visualization of flatness near the solutions obtained by Adam and DualAdam. Finally, we conduct ablation studies to analyze the impact of the switching rates, switching mechanisms, and learning rate schedulers on DualAdam's performance. The detailed settings of all the simulations and experiments are provided in Supplementary File. Additionally, we conduct experiments to investigate the impact of different learning rate schedulers on the performance of DualAdam, which is also included in Supplementary File.

\subsection{Numerical Simulations on 2-Parameter Loss Landscapes}

\begin{figure}[h!]
    \centering
  \subfloat[Custom Function]{
    \includegraphics[width=0.48\linewidth]{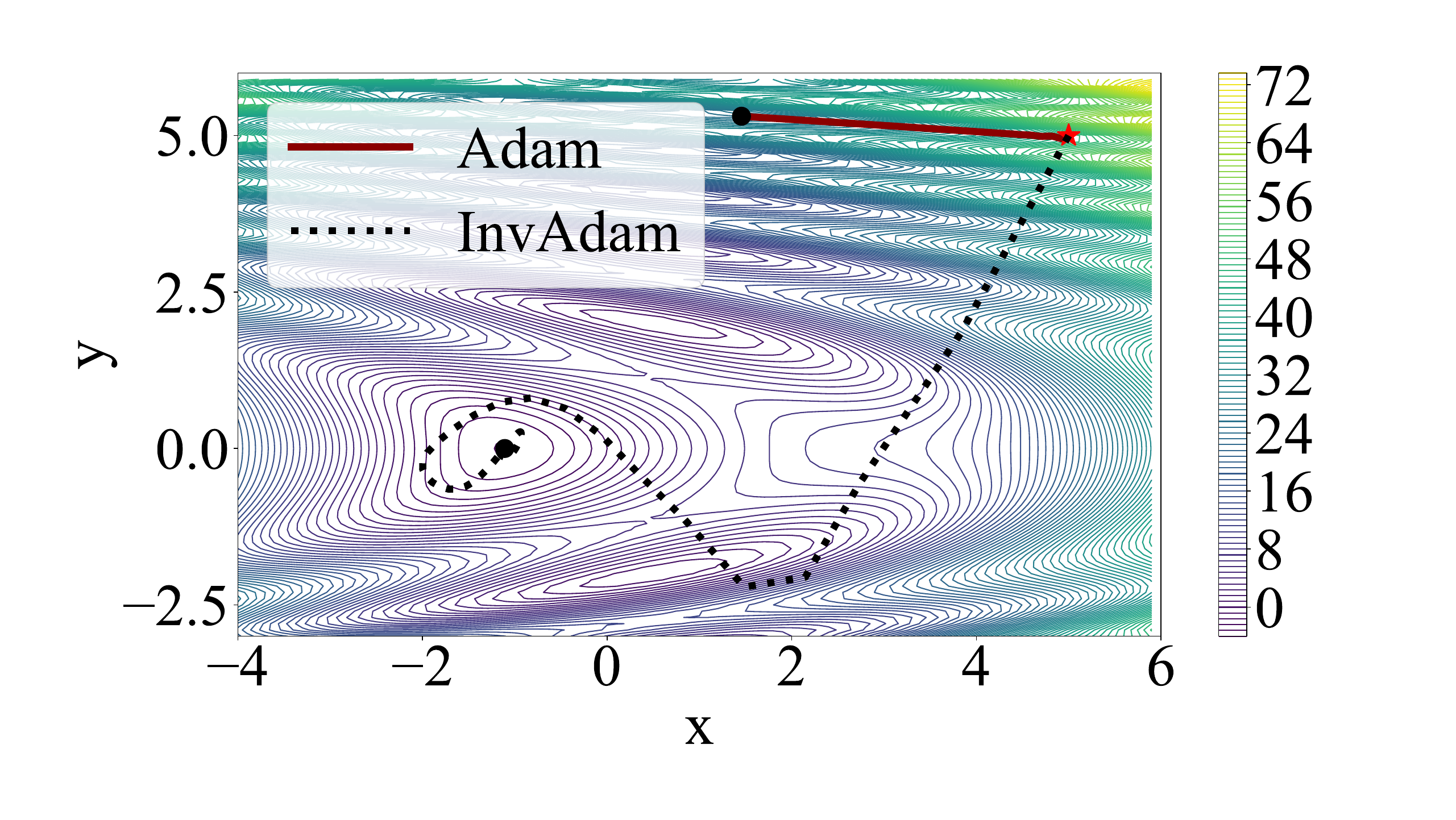}
  }
	\subfloat[Eggholder Function]{
    \includegraphics[width=0.48\linewidth]{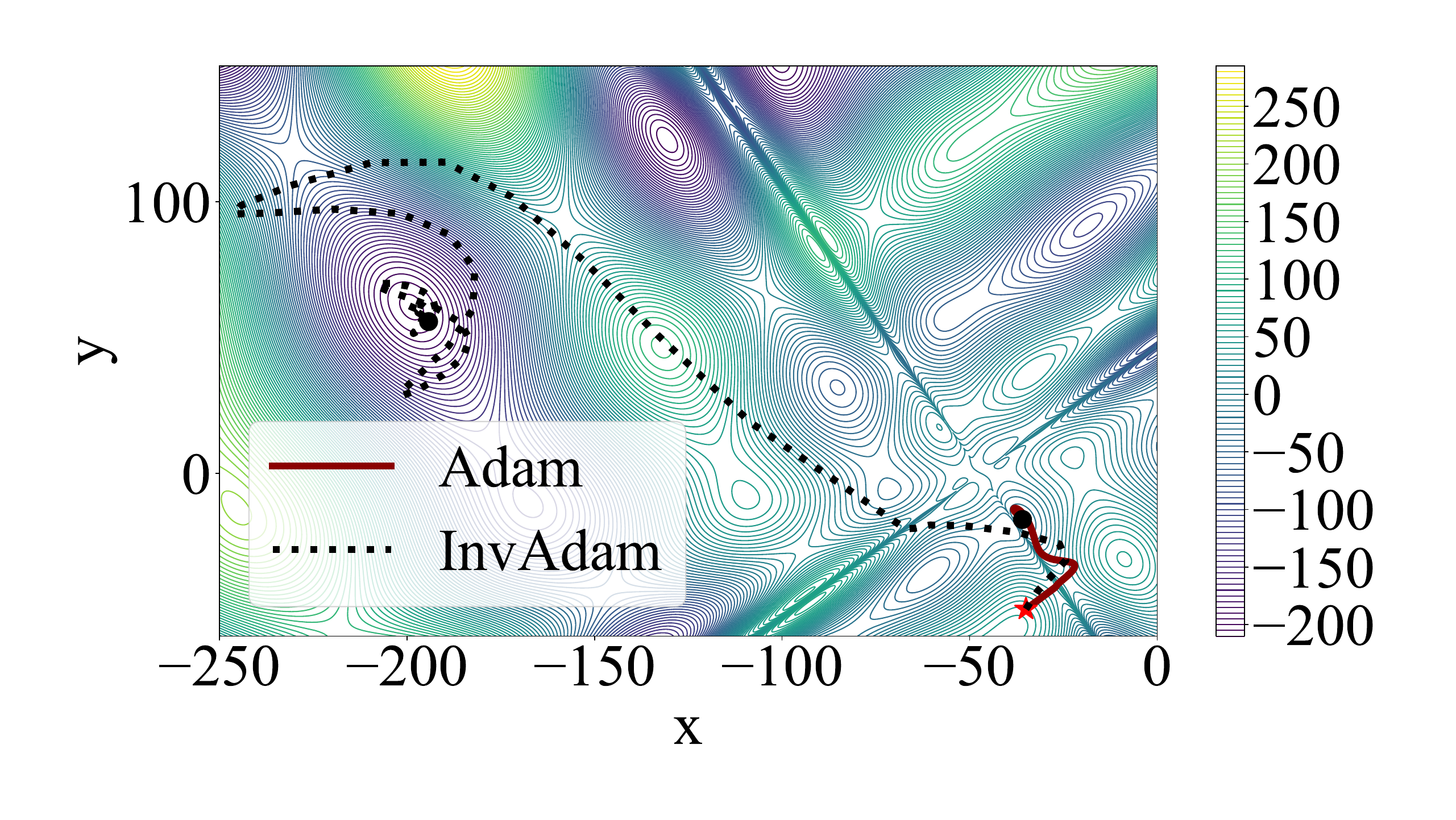}
  }
    \caption{Optimization trajectories of InvAdam and Adam on 2-parameter loss landscapes. The red stars represent the start points, and the black circles represent the end points.}
    \label{fig_sim}
\end{figure}

In this section, we conduct numerical simulations on 2-parameter loss landscapes to demonstrate that InvAdam exhibits a better ability to escape sharp minima than Adam. The loss landscapes we use include a loss landscape determined by a custom function and a loss landscape determined by the Eggholder function \cite{Jung2024Bypassing}. As shown in Fig. \ref{fig_sim}, it is evident that Adam quickly gets trapped in a sharp minimum, whereas InvAdam explores the loss landscape more thoroughly and eventually converges to a flat minimum. These trajectories demonstrate that InvAdam has superior ability to find flat minima over Adam. However, the ablation studies on the switching rate indicate that the sole use of InvAdam may lead to non-convergence. This is why DualAdam uses InvAdam in the early stage of training and switches to Adam as the training progresses.

\subsection{Image Classification on CIFAR-10 and CIFAR-100}

\begin{figure}[h!]
    \centering
  \subfloat[ResNet-18]{
    \includegraphics[width=0.48\linewidth]{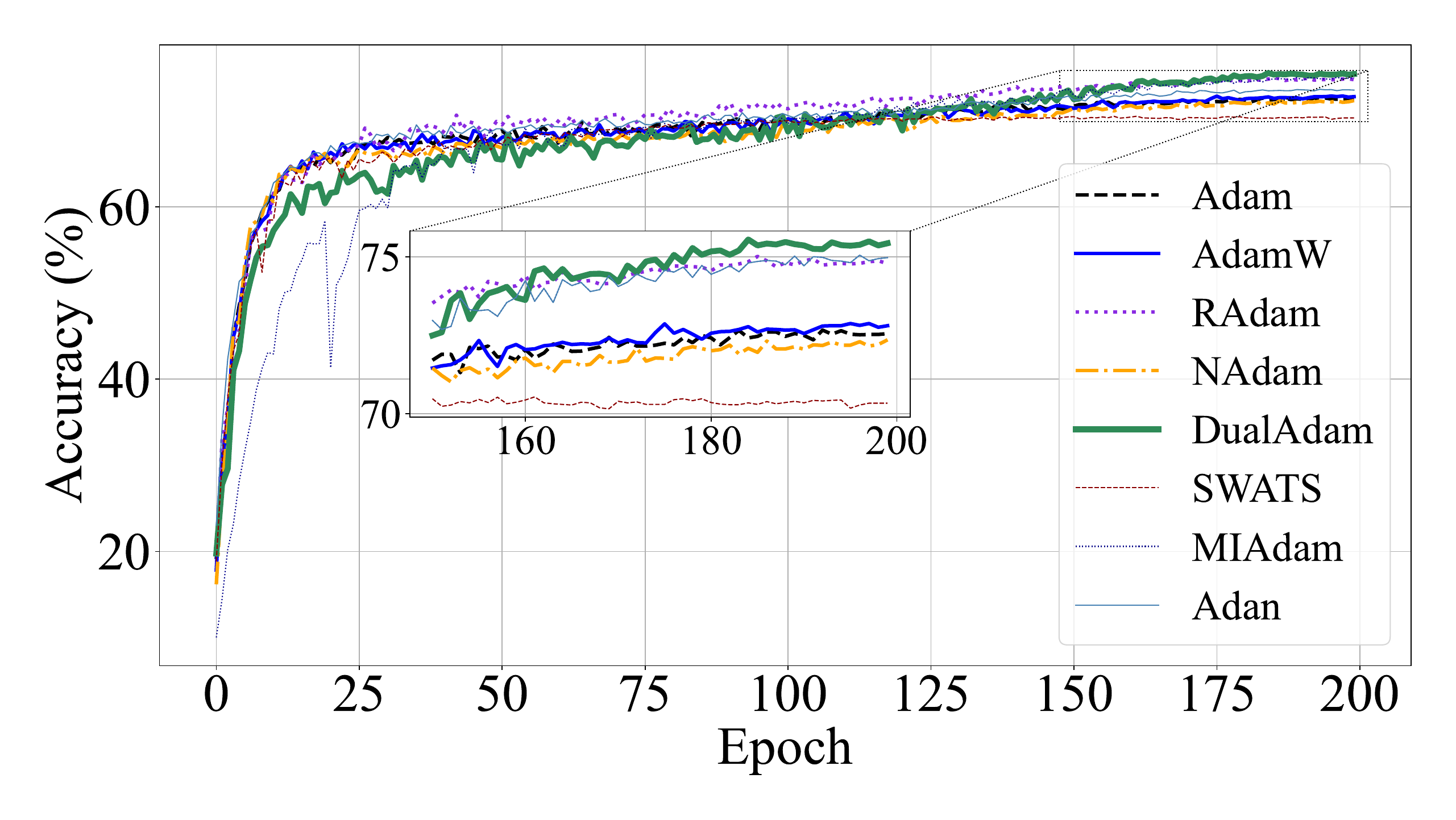}
  }
	\subfloat[ViT-Small-4]{
    \includegraphics[width=0.48\linewidth]{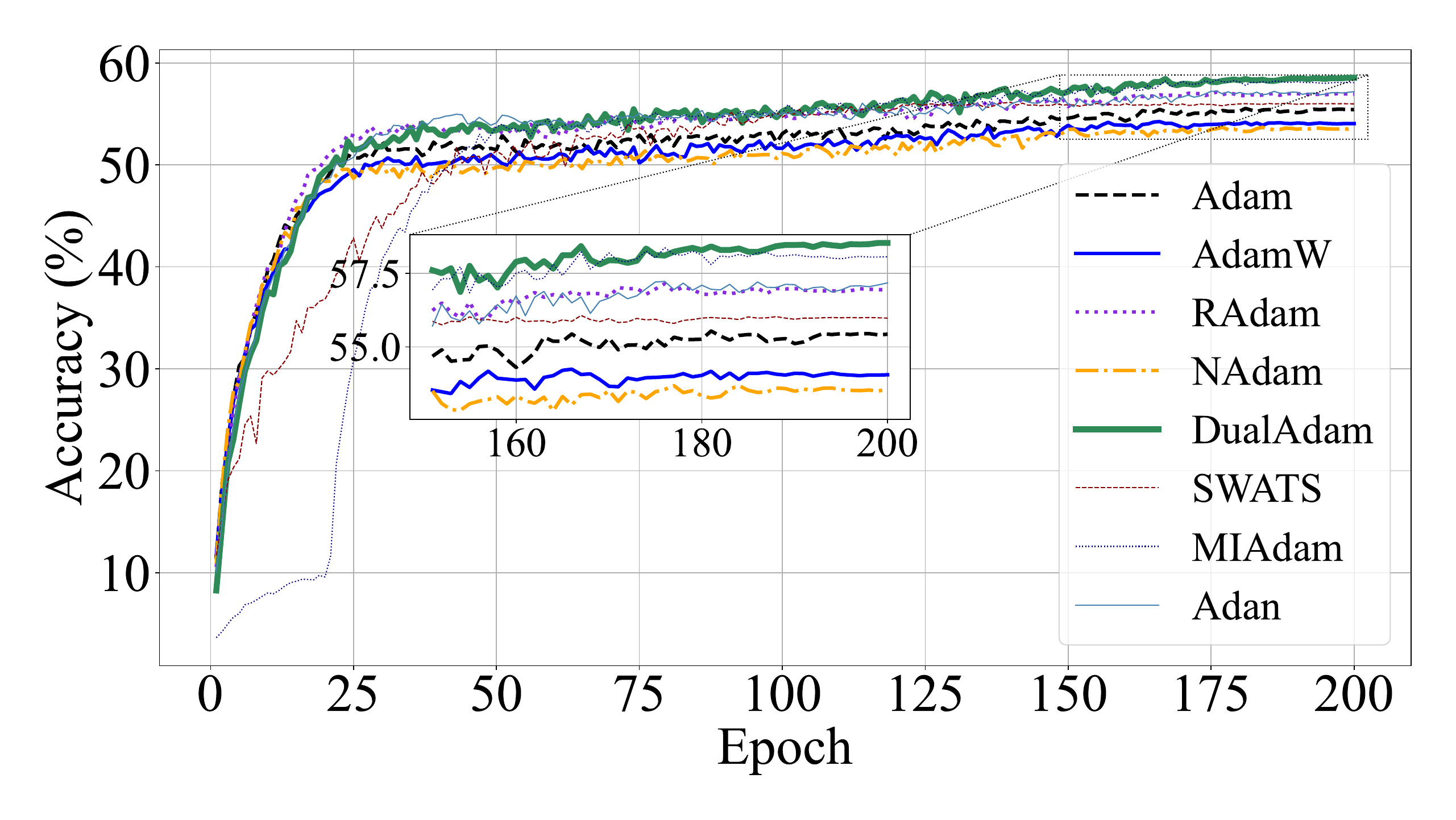}
  }
    \caption{Test accuracies over epochs on CIFAR-100.}
    \label{fig_ac_cf100}
\end{figure}

We conduct experiments on the CIFAR-10 and CIFAR-100 datasets. The training data preprocessing involves randomly cropping the images to a size of $32 \times 32$ with a 4-pixel padding and performing random horizontal flips to augment the dataset. Subsequently, the images are normalized using mean and standard deviation values calculated from the dataset. Figure \ref{fig_ac_cf100} shows the test accuracy over epochs for different optimizers using ResNet-18 and ViT-Small-4 on CIFAR-100. It demonstrates that DualAdam achieves higher test accuracy than Adam and its variants after the training is complete. Each optimizer is run three times on both CIFAR-10 and CIFAR-100 with different models, and the mean and standard deviation of test accuracies are reported in Table \ref{tb_cf}, with the training time for each single run. As shown, DualAdam outperforms Adam and its state-of-the-art variants in terms of generalization, with nearly the same training time as Adam's. The results validate that DualAdam effectively balances convergence and generalization by combining the update mechanisms of Adam and InvAdam.

\begin{table*}[h!] 
\centering
\caption{Top-1 test accuracies (mean±std) and training times on CIFAR-10 and CIFAR-100}
\begin{tabular}{cccccc}
\toprule
\multirow{2}{*}{Model} & \multirow{2}{*}{Optimizer} & \multicolumn{2}{c}{CIFAR-10} & \multicolumn{2}{c}{CIFAR-100} \\
\cmidrule(r){3-4} \cmidrule(r){5-6}
& & Test Accuracy (\%) & Training Time & Test Accuracy (\%)  & Training Time \\
\midrule
\multirow {8}{*}{ResNet-18} 
& Adam \cite{kingma2014adam} & $93.66_{\pm0.17}$ & 36min40s  & $72.56_{\pm0.08}$  & 36min37s \\
& AdamW \cite{loshchilov2017decoupled} & $94.09_{\pm0.04}$ & 36min43s  & $72.81_{\pm0.10}$ & 37min10s \\
& SWATS \cite{keskar2017improving} & $92.96_{\pm1.04}$ & 43min27s  &  $71.40_{\pm1.50}$& 44min04s \\
& NAdam \cite{dozat2016incorporating} & $94.01_{\pm0.02}$ & 37min12s  & $72.06_{\pm0.43}$ & 37min35s \\
& RAdam \cite{liu2019variance} & $94.11_{\pm0.01}$ & 37min11s  & $74.79_{\pm0.06}$ & 36min50s \\
& Adan \cite{xie2024adan} & $94.07_{\pm0.07}$ & 37min11s    & $73.57_{\pm0.12}$  & 36min41s  \\
& MIAdam \cite{jin2025nong} & $94.25_{\pm0.15}$ & 37min18s  & $74.98_{\pm0.03}$  & 38min23s \\
& DualAdam (ours) & $\mathbf{94.99_{\pm0.13}}$ & 37min06s  & $\mathbf{75.29_{\pm0.21}}$ & 36min54s \\
\midrule
\multirow {8}{*}{ResNet-50} 
& Adam \cite{kingma2014adam} & $94.04_{\pm 0.12}$ & 101min12s   & $74.53_{\pm 0.50}$  & 100min18s  \\
& AdamW \cite{loshchilov2017decoupled} & $94.30_{\pm 0.02}$  & 103min01s  & $75.40_{\pm 0.31}$  & 102min18s  \\
& SWATS \cite{keskar2017improving} & $90.99_{\pm 1.56}$  & 110min45s & $71.87_{\pm 0.67}$  & 113min20s \\
& NAdam \cite{dozat2016incorporating} & $94.26_{\pm 0.28}$  & 103min19s & $74.74_{\pm 0.11}$ & 104min25s  \\
& RAdam \cite{liu2019variance} & $94.76_{\pm 0.07}$  & 101min07s  & $76.25_{\pm 0.17}$  & 102min13s  \\
& Adan \cite{xie2024adan}  & $94.62_{\pm 0.11}$  & 100min32s  & $75.96_{\pm 0.05}$  & 101min09s  \\
& MIAdam \cite{jin2025nong}& $94.45_{\pm 0.08}$  & 101min19s  & $76.70_{\pm 0.07}$ & 102min30s \\
& DualAdam (ours) & $\mathbf{95.27_{\pm0.08}}$  & 101min34s  & $\mathbf{76.84_{\pm 0.16}}$  & 102min15s \\
\midrule
\multirow {8}{*}{VGG-16} 
& Adam \cite{kingma2014adam}  & $92.98_{\pm0.15}$ & 29min03s & $68.66_{\pm0.13}$  & 28min31s \\
& AdamW \cite{loshchilov2017decoupled} & $92.99_{\pm0.19}$ & 29min15s & $68.69_{\pm0.12}$  & 29min12s \\
& SWATS \cite{keskar2017improving} & $91.41_{\pm1.49}$ & 35min21s & $65.56_{\pm1.40}$  & 33min23s\\
& NAdam \cite{dozat2016incorporating} & $92.98_{\pm0.02}$ & 30min15s & $69.85_{\pm0.07}$  & 30min03s\\
& RAdam \cite{liu2019variance} & $93.36_{\pm0.25}$ & 31min08s & $70.49_{\pm0.17}$  & 31min22s\\
& Adan \cite{xie2024adan}  & $92.88_{\pm0.03}$ & 32min01s & $70.04_{\pm0.16}$  & 31min06s \\
& MIAdam \cite{jin2025nong}& $93.13_{\pm0.11}$ & 30min33s & $71.92_{\pm0.11}$  & 30min03s \\
& DualAdam (ours) & $\mathbf{93.93_{\pm0.20}}$ & 30min15s  & $\mathbf{72.66_{\pm0.21}}$  & 30min18s \\
\midrule
\multirow {8}{*}{ViT-Small-4} 
& Adam \cite{kingma2014adam}  & $83.30_{\pm0.42}$ & 66min40s  & $55.43_{\pm0.09}$  & 67min10s \\
& AdamW \cite{loshchilov2017decoupled} & $83.16_{\pm0.22}$ & 67min12s & $54.05_{\pm0.13}$  & 67min08s\\
& SWATS \cite{keskar2017improving} & $79.29_{\pm1.10}$ & 75min13s & $55.95_{\pm0.03}$  & 72min43s\\
& NAdam \cite{dozat2016incorporating} & $83.50_{\pm0.24}$ & 68min14s & $53.55_{\pm0.03}$  & 69min21s\\
& RAdam \cite{liu2019variance} & $84.15_{\pm0.05}$ & 68min06s & $56.95_{\pm0.33}$  & 68min12s\\
& Adan \cite{xie2024adan}  & $83.71_{\pm0.04}$ & 66min15s & $57.16_{\pm0.07}$  & 70min45s\\
& MIAdam \cite{jin2025nong} & $83.45_{\pm0.18}$& 67min32s & $58.01_{\pm0.11}$  & 71min12s\\
& DualAdam (ours) & $\mathbf{85.40_{\pm0.17}}$ & 68min15s & $\mathbf{58.54_{\pm0.24}}$ &69min13s\\

\bottomrule
\end{tabular}
\label{tb_cf}
\end{table*}

\subsection{Image Classification on Tiny ImageNet}

We conduct experiments on Tiny ImageNet, a subset of ImageNet-1k comprising 200 classes with 500 training images, 50 validation images, and 50 test images per class. It is worth noting that Tiny ImageNet reduces the size of the images in ImageNet-1k to $64 \times 64$ (most of the images in ImageNet-1k are larger than $224 \times 224$), which makes them appear more blurry compared to the images in ImageNet-1k and therefore more difficult to classify. The training data preprocessing involves several steps to augment and standardize the images in Tiny ImageNet. Initially, a random horizontal flip is performed to enhance the variability in the training data. After that, the images are normalized using mean and standard deviation values calculated from the dataset. Each optimizer is run three times on Tiny ImageNet with different models, and the mean and standard deviation of test accuracies are reported in Table \ref{tb_in}. As shown, DualAdam also outperforms Adam and some of its variants on Tiny ImageNet.

\begin{table}[h!]
\centering
\caption{Test accuracies (mean±std) on Tiny ImageNet}
\begin{tabular}{cccc}
\toprule
 Model & Optimizer & \multicolumn{1}{c}{Test accuracy (\%)} & Traning Time \\
\midrule
\multirow{8}{*}{ResNet-18}  
& Adam \cite{kingma2014adam}  & $50.98_{\pm0.42}$ & 537min24s \\                      
& AdamW \cite{loshchilov2017decoupled} & $51.20_{\pm0.28}$ & 536min18s \\   
& SWATS \cite{keskar2017improving}     & $49.55_{\pm0.45}$ & 601min18s\\                        
& NAdam \cite{dozat2016incorporating}  & $51.58_{\pm0.08}$ & 538min13s\\         
& RAdam \cite{liu2019variance}         & $55.04_{\pm0.32}$ & 537min06s\\
& Adan  \cite{xie2024adan}             & $51.97_{\pm0.21}$ & 540min18s\\
& MIAdam  \cite{jin2025nong}           & $53.06_{\pm0.34}$ & 536min14s\\
& DualAdam (ours)                      & $\mathbf{57.19_{\pm0.21}}$ & 538min45s \\
\midrule
\multirow{8}{*}{VGG-16}    
 & Adam \cite{kingma2014adam}           & $48.57_{\pm0.10}$ & 402min06s \\
 & AdamW \cite{loshchilov2017decoupled} & $50.53_{\pm0.14}$ & 405min21s\\
 & SWATS \cite{keskar2017improving}     & $44.14_{\pm0.64}$ & 435min45s\\   
 & NAdam \cite{dozat2016incorporating}  & $48.80_{\pm0.08}$ & 406min18s\\
 & RAdam \cite{liu2019variance}         & $53.01_{\pm0.08}$ & 401min29s\\
 & Adan  \cite{xie2024adan}             & $51.52_{\pm0.12}$ & 402min17s\\
 & MIAdam\cite{jin2025nong}             & $53.11_{\pm0.11}$ & 402min25s\\
 & DualAdam (ours)                      & $\mathbf{54.42_{\pm0.11}}$ & 403min33s\\
\midrule
\multirow{8}{*}{ViT-Small-8} 
 & Adam \cite{kingma2014adam}           & $30.72_{\pm0.21}$ & 569min33s\\
 & AdamW \cite{loshchilov2017decoupled} & $31.20_{\pm0.13}$ & 567min40s\\
 & SWATS \cite{keskar2017improving}     & $28.80_{\pm0.11}$ & 586min32s\\
 & NAdam \cite{dozat2016incorporating}  & $28.56_{\pm0.03}$ & 572min08s\\
 & RAdam \cite{liu2019variance}         & $34.22_{\pm0.32}$ & 570min18s\\
 & Adan  \cite{xie2024adan}             & $33.53_{\pm0.09}$ & 566min51s\\
 & MIAdam\cite{jin2025nong}             & $35.85_{\pm0.07}$ & 572min18s\\
 & DualAdam (ours)                      & $\mathbf{37.90_{\pm0.22}}$ & 568min19s\\
\bottomrule
\end{tabular}
\label{tb_in}
\end{table}

\subsection{Image Classification on ImageNet-1k}

We conduct experiments on ImageNet-1k, a large-scale image classification dataset comprising 1,000 classes with 1.28 million training images and 50,000 testing images. The training data preprocessing involves several steps to augment and standardize the images. Initially, the images are randomly cropped to a size of $224 \times 224$. After that, a random horizontal flip is performed to enhance variability in the training data. Then, the images are normalized using mean and standard deviation values calculated from the dataset. The test accuracies are reported in Table \ref{tb_imagent}. As shown, DualAdam outperforms Adam on a relatively large dataset.

\begin{table}[h!]
\centering
\caption{Test accuracies on ImageNet-1k}
\begin{tabular}{ccc}
\toprule
 Model & Optimizer & \multicolumn{1}{c}{Test Accuracy (\%)} \\
\midrule
\multirow{3}{*}{ResNet-18}
                           & Adam \cite{kingma2014adam}           & $69.55$ \\
                           & AdamW \cite{loshchilov2017decoupled} & $71.29$ \\
						   & DualAdam (ours)                      & $\mathbf{72.61}$\\
\midrule
\multirow{3}{*}{ResNet-50} & Adam  \cite{kingma2014adam}          & $70.10$\\
                           & AdamW \cite{loshchilov2017decoupled} & $71.23$\\
						   & DualAdam (ours)                      & $\mathbf{73.32}$ \\
\bottomrule
\end{tabular}
\label{tb_imagent}
\end{table}

\subsection{Fine-Tuning on Large Language Model}

\begin{figure*}[h!]
    \centering
  \subfloat[Training Loss]{
    \includegraphics[width=0.3\linewidth]{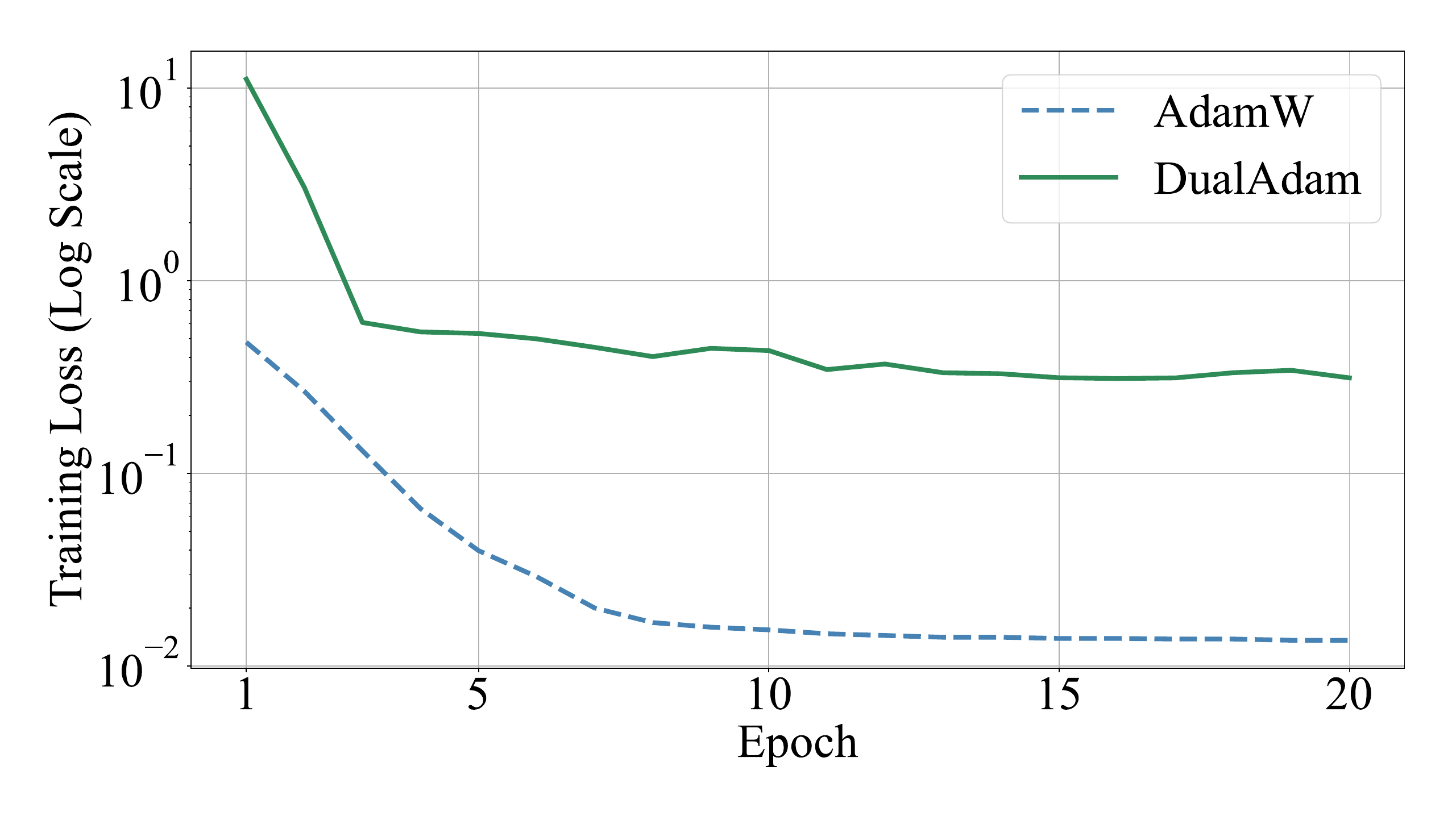}
  }
	\subfloat[Validation Perplexity]{
    \includegraphics[width=0.3\linewidth]{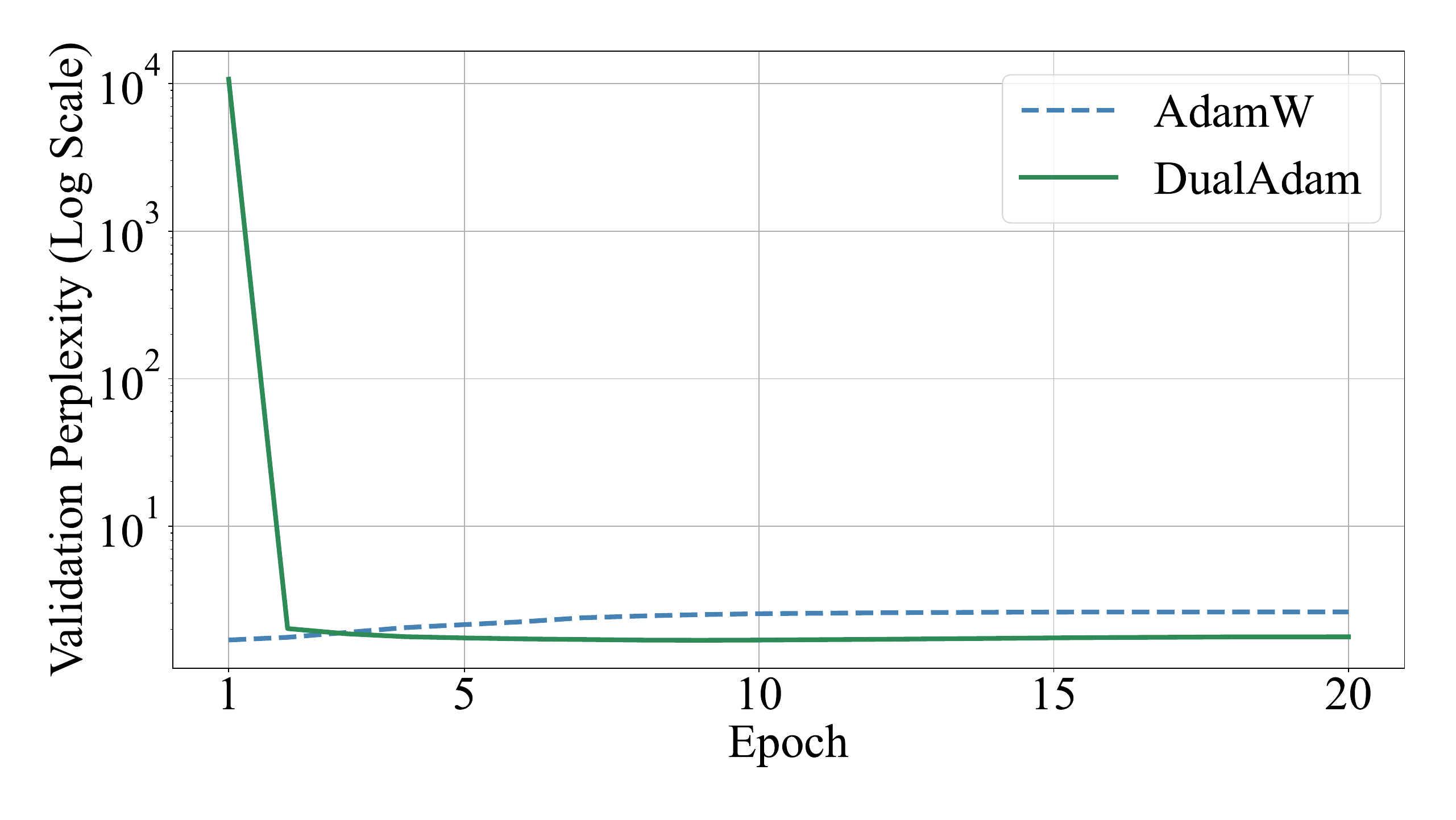}
  }
  \subfloat[Generalization Gap]{
    \includegraphics[width=0.3\linewidth]{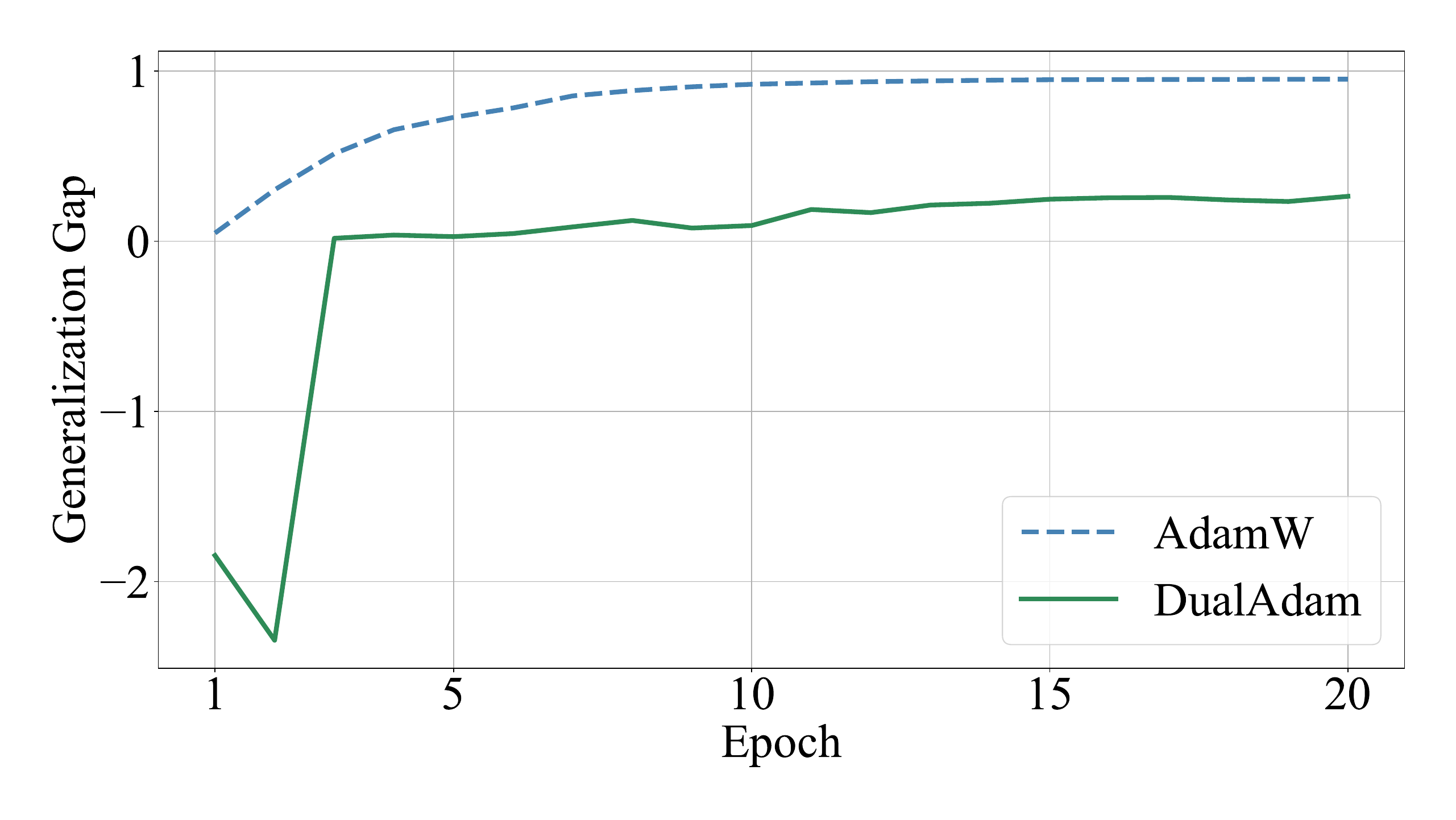}
  }
    \caption{Comparisons of training loss, validation perplexity, and generalization gap of DualAdam and AdamW on the fine-tuning of OpenPangu-1B.}
    \label{fig_openpangu}
\end{figure*}

To further evaluate the scalability and versatility of DualAdam on large-scale parameters and language modeling tasks, we conduct fine-tuning experiments on an large language model (LLM). We employ the OpenPangu-Embedded-1B model \cite{chen2025pangu}, a LLM with a billion parameters. The model is fine-tuned on the Alpaca-GPT4-CN dataset\footnote{https://huggingface.co/datasets/surogate/alpaca-gpt4-data-zh}, which consists of high-quality Chinese instruction-following data. We compare DualAdam against AdamW, which is the default optimizer for training LLMs. Fig. \ref{fig_openpangu} illustrates the comparison results. As shown, AdamW achieves a significantly lower training loss compared to DualAdam. However, despite the higher training loss, DualAdam achieves a remarkably lower and more stable validation perplexity (PPL) compared to AdamW. As training progresses, AdamW's PPL begins to rise, a classic sign of overfitting. In contrast, DualAdam's PPL continues to decrease or remains flat, demonstrating robust generalization. Moreover, the advantage of DualAdam is evident in the generalization gap, which is the difference between the validation loss and training loss. AdamW exhibits a rapidly increasing generalization gap, confirming severe overfitting. DualAdam maintains a generalization gap near zero, empirically validating our theoretical claim that DualAdam guides the parameters toward flat minima, which are robust to data variations. These results on OpenPangu-Embedded-1B confirm that DualAdam's benefits extend beyond computer vision to natural language processing, effectively handling the overfitting challenges inherent in LLM fine-tuning.

\subsection{Comparison of the Hessian Matrix's Information}

\begin{figure}[h!]
    \centering
  \subfloat[Adam]{
    \includegraphics[width=0.47\linewidth]{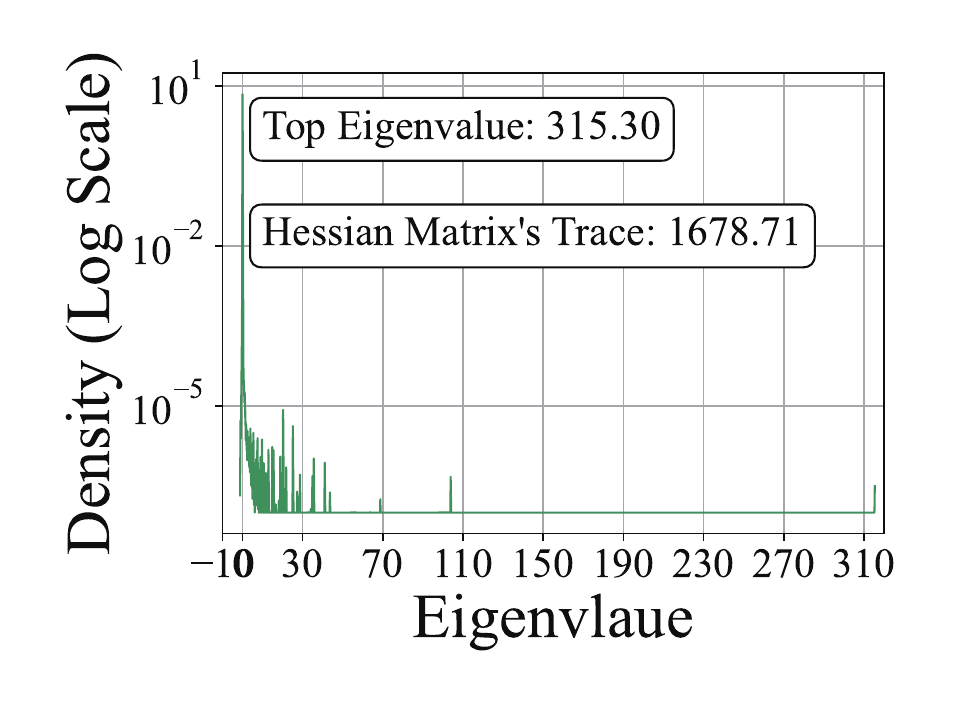}
  }
	\subfloat[DualAdam]{
    \includegraphics[width=0.47\linewidth]{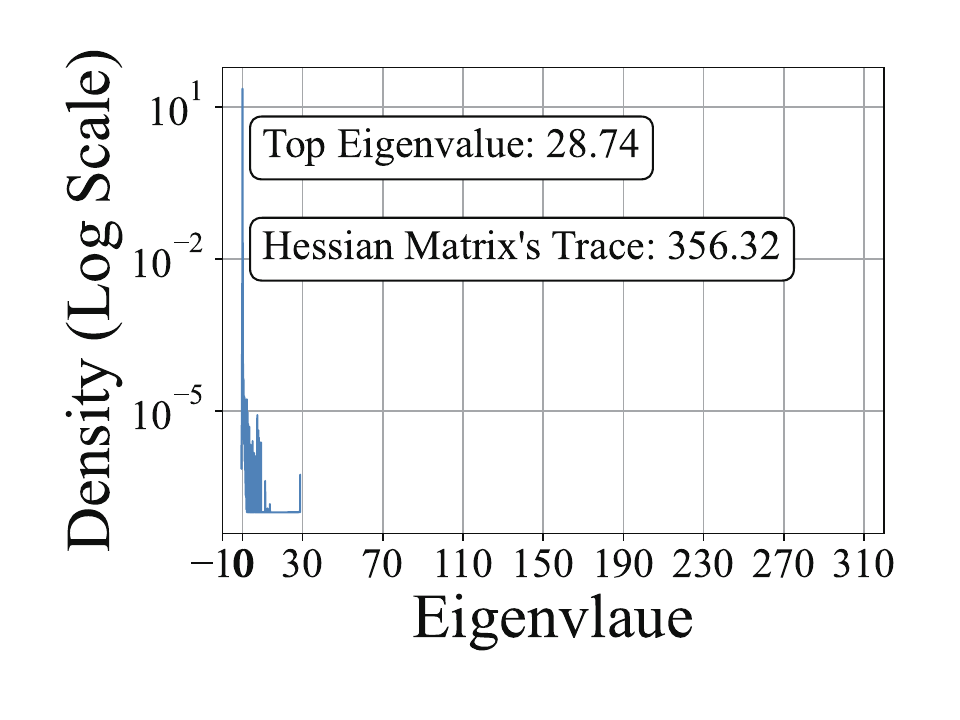}
  }
    \caption{Comparisons of top Hessian matrix's eigenvalues, Hessian matrix's traces, and Hessian matrix's eigenvalue densities of loss landscapes on the CIFAR-100 dataset using ResNet18.}
    \label{fig_hessian}
\end{figure}

We employ the method in \cite{yao2020pyhessian} to compute the eigenvalues of the Hessian matrix for the solutions obtained by Adam and DualAdam using ResNet-18 on CIFAR-100. As shown in Fig. \ref{fig_hessian}, the Hessian eigenvalues of model parameters optimized by DualAdam are more concentrated around zero compared to those optimized by Adam, with smaller maximum eigenvalues and trace. This indicates that the model parameters optimized by DualAdma reside in a flatter basin of the loss landscape compared to those optimized by Adam.

\subsection{Visualizing Loss Landscape Flatness}
\begin{figure}[h!]
  \centering
	\includegraphics[width=0.8\linewidth]{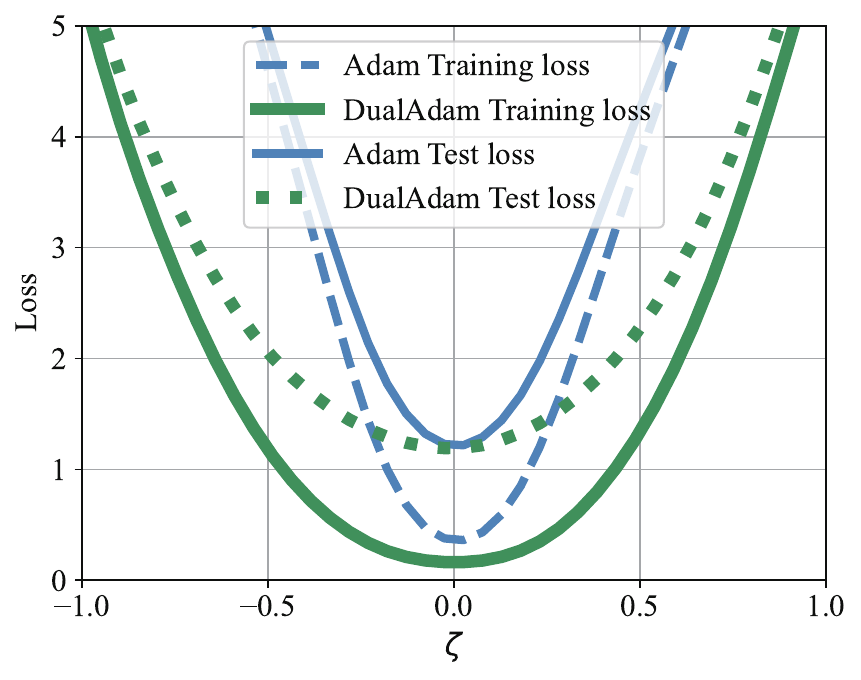}
  \caption{Visualization of loss landscapes of Adam and DualAdam. The loss is defined as $L(\theta^\star + \zeta l)$, where $\zeta$ is a scalar that scales the random vector $l$ drawn from a Gaussian distribution to perturb the optimal parameter $\theta^\star$.}
    \label{fig_lsld}
\end{figure}

To measure the flatness of the loss landscape near the solutions obtained by the different optimizers, we use the method in \cite{visualloss} for visualization. We present a 1D visualization of solutions achieved by Adam and DualAdam using ResNet-18 on CIFAR-10. As illustrated in Fig. \ref{fig_lsld}, it is evident that DualAdam obtains a flatter solution than Adam, indicating its better generalization performance \cite{yang2023stochastic}.

\begin{table}[h!]
\centering
\caption{Training losses (mean±std) and validation accuracies (mean±std) of DualAdam on CIFAR-100 using ResNet-18 with different switching rates}
\begin{tabular}{ccc}
\toprule
Switching Rate & Traning Loss &Validation Accuracy (\%) \\
\midrule
0                  & $4.58_{\pm 0.01}$  & $2.0_{\pm 0.04}$ \\
\midrule
$1 \times 10^{-5}$ & $5.4_{\pm 0.35} \times 10^{-4}$  & $69.04_{\pm 0.08}$ \\
\midrule
$2 \times 10^{-5}$ & $5.5_{\pm 0.77} \times 10^{-4}$  & $70.30_{\pm 0.26}$ \\
\midrule
$3 \times 10^{-5}$ & $4.8_{\pm 0.07} \times 10^{-4}$  & $71.04_{\pm0.08}$\\	
\midrule
$4 \times 10^{-5}$ & $5.1_{\pm 0.35} \times 10^{-4}$  & $71.49_{\pm0.21}$\\
\midrule
$5 \times 10^{-5}$ & $4.8_{\pm 0.14} \times 10^{-4}$  & $71.52_{\pm0.28}$\\					  
\midrule
$6 \times 10^{-5}$ & $4.5_{\pm 0.42} \times 10^{-4}$ & $71.59_{\pm 0.49}$\\
\midrule
$7 \times 10^{-5}$ & $4.7_{\pm 0.07} \times 10^{-4}$ & $71.38_{\pm 0.02}$\\
\midrule
$8 \times 10^{-5}$ & $4.5_{\pm 0.06} \times 10^{-4}$ & $71.79_{\pm 0.28}$\\
\midrule
$9 \times 10^{-5}$ & $4.3_{\pm 0.14} \times 10^{-4}$ & $71.50_{\pm 0.33}$\\
\midrule
$1 \times 10^{-4}$ & $4.9_{\pm 0.07} \times 10^{-4}$ & $71.48_{\pm 0.28}$\\
\bottomrule
\end{tabular}
\label{tb_ablation_switch}
\end{table}

\begin{table}[h!]
\centering
\caption{Test accuracies (mean±std) of DualAdam on CIFAR-100 using ResNet-18 with different switching mechanisms}
\begin{tabular}{ccc}
\toprule
 \multirow{5}{*}{Linear Switching} & Switching Rate &Test Accuracy (\%) \\
\cmidrule(l){2-3}
                                   &  $5 \times 10^{-5}$  &$75.30_{\pm 0.34}$  \\
								   &  $8 \times 10^{-5}$  &$75.29_{\pm 0.21}$\\
								   &  $1 \times 10^{-4}$  &$75.13_{\pm 0.16}$\\
\midrule
\multirow{5}{*}{Exponential Switching}&Exponential Base & Test Accuracy (\%)  \\
\cmidrule(l){2-3}
                                      &  0.8       & $71.45_{\pm0.47}$\\	
									  &  0.9       & $72.82_{\pm0.09}$\\
									  &  0.99      & $73.33_{\pm0.39}$\\					  
\midrule
\multirow{5}{*}{Fixed Epoch Switching} & Switching Epoch & Test Accuracy (\%) \\
\cmidrule(l){2-3}
                                       &  10 &  $71.33_{\pm0.56}$ \\
								       &  30 &  $69.03_{\pm0.31}$ \\
								       &  50 &  $66.96_{\pm0.46}$ \\
\bottomrule
\end{tabular}
\label{tb_ablation_switch}
\end{table}

\subsection{Ablation Study}

We conduct ablation studies to investigate the impact of the switching rate and switching mechanisms on the performance of DualAdam. The experiments are conducted using ResNet-18 on CIFAR-100. The detailed settings of the experiments are provided in Supplementary File.

\subsubsection{Switching Rate}

We conduct ablation studies to investigate the impact of switching rate $\xi$ on the performance of DualAdam. $\xi$ determines how quickly DualAdam transitions from InvAdam to Adam during training. Small $\xi$ means slow transition, while large $\xi$ means fast one. We perform experiments by using ResNet-18 on the CIFAR-100 training set with different values of $\xi$. 80\% of the training set is used for training, and the remaining 20\% is used for validation. The experiments are run three times, and the average values and standard deviations of the training losses and top-1 validation accuracies of the last epoch are reported in Table \ref{tb_ablation_switch}. As shown, when $\xi$ is set to $0$, i.e., InvAdam is used solely, the model's training may not converge. When $\xi$ is set to $8 \times 10^{-5}$, DualAdam achieves the best test performance. Therefore, we set $\xi$ to $8 \times 10^{-5}$ in the image classification experiments. Moreover, if $\xi$ is too small or too large, the performance of DualAdam degrades. This indicates that an appropriate switching rate is crucial for balancing exploration and convergence in DualAdam.

\subsubsection{Switching Mechanisms}
We conduct ablation studies to investigate the impact of different switching mechanisms on DualAdam's performance. We compare three switching mechanisms: linear, exponential, and fixed epoch ones. As shown in Table \ref{tb_ablation_switch}, linear switching achieves the best performance among them. Exponential switching performs slightly worse than linear switching, while fixed epoch switching performs the worst. This indicates that a gradual transition from InvAdam to Adam is more effective than an abrupt transition.

\section{Conclusions}
We have proposed a new optimizer, named InvAdam, and an enhanced version of it, called DualAdam. While InvAdam has a better ability to escape sharp minima than Adam theoretically, it faces potential challenges in achieving convergence. Therefore, we have proposed DualAdam, which combines both. The linear switching between them can effectively balance generalization and convergence. Then, we have provided a theoretical analysis to mathematically demonstrate that InvAdam has a better ability to escape sharp minima than Adam. Additionally, we have validated through extensive experiments that DualAdam has a better generalization performance than Adam and some of its state-of-the-art variants on both image classification and LLM fine-tuning tasks. In the future, we plan to explore more effective switching mechanisms and investigate the performance of the combination of InvAdam and other optimizers, such as SGD.
 
\bibliographystyle{IEEEtran}
\bibliography{references.bib}

\begin{IEEEbiography}[{\includegraphics[width=1in,height=1.25in,clip,keepaspectratio]{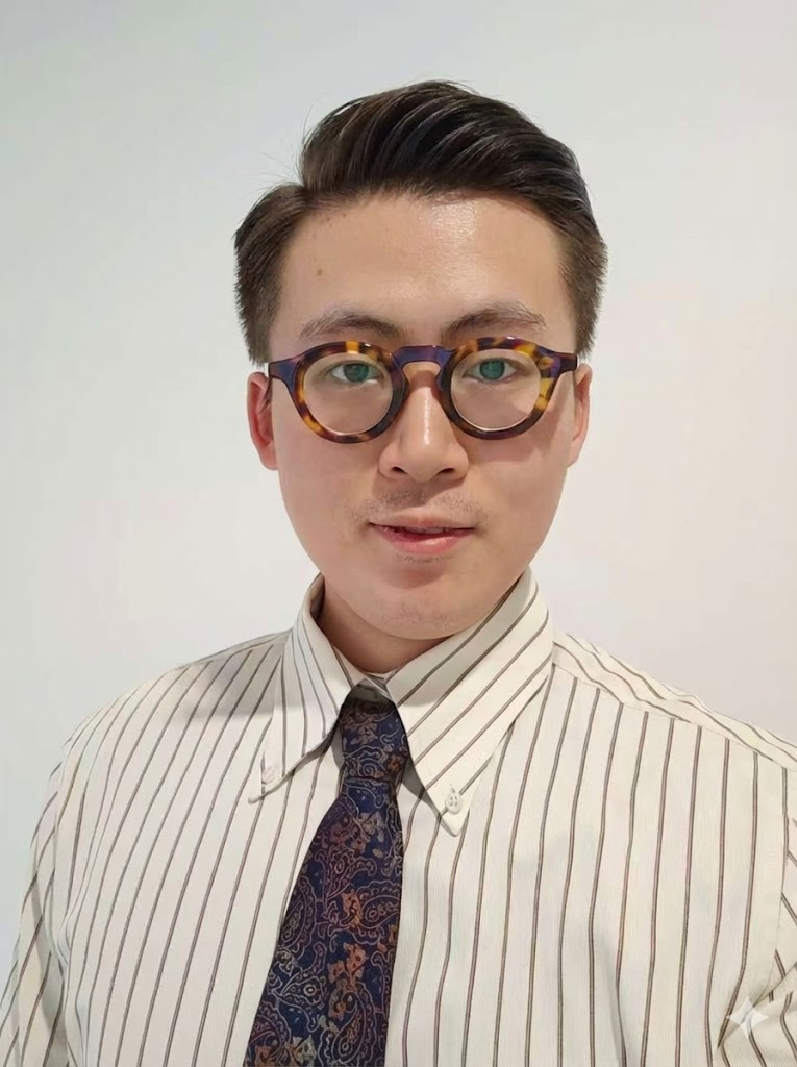}}]{Tao Shi} received the B.E. degree in software engineering from Xiamen University, Xiamen, China, in 2022. He is currently pursuing an M.E. degree in computer science and technology with the School of Information Science and Engineering, Lanzhou University, Lanzhou, China.

His research interests include optimization and deep learning.
\end{IEEEbiography}

\begin{IEEEbiography}[{\includegraphics[width=1in,height=1.25in,clip,keepaspectratio]{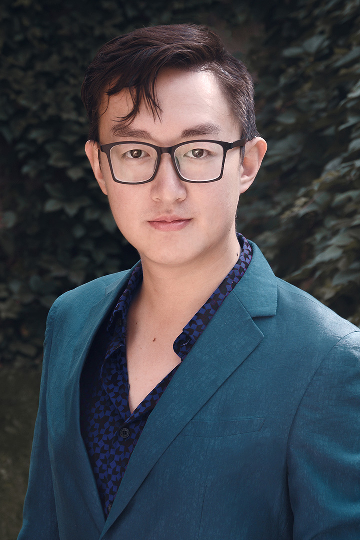}}]{Liangming Chen}  received a B.S. degree in physics from Wuhan University, Wuhan, China, in 2017, an ME degree in computer technology from Lanzhou University, Lanzhou, China, in 2021, and a Ph.D. degree in computer software and theory with the University of Chinese Academy of Sciences, Beijing, China, in 2025. 
	
His main research interests include structures and training algorithms of deep neural networks.
\end{IEEEbiography}

\begin{IEEEbiography}[{\includegraphics[width=1in,height=1.25in,clip,keepaspectratio]{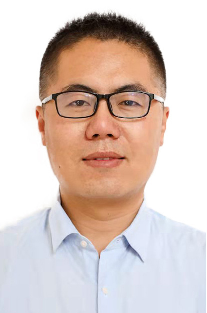}}]{Long Jin} (Senior Member, IEEE) received the B.E. degree in automation and the Ph.D. degree in information and communication engineering from Sun Yat-sen University, Guangzhou, China, in 2011 and 2016, respectively. He was a Postdoctoral Fellow with the Department of Computing, The Hong Kong Polytechnic University, Hong Kong, from 2016 to 2017. In 2017, he joined the School of Information Science and Engineering, Lanzhou University, Lanzhou, China, as a Professor of Computer Science and Engineering. He is currently serving as the Associate Editor for the IEEE Transactions on Industrial Electronics, IEEE/CAA Journal of Automatica Sinica, and Neural Networks. His current research interests include neural networks, robotics, optimization, and intelligent computing.
\end{IEEEbiography}

\begin{IEEEbiography}[{\includegraphics[width=1in,height=1.25in,clip,keepaspectratio]{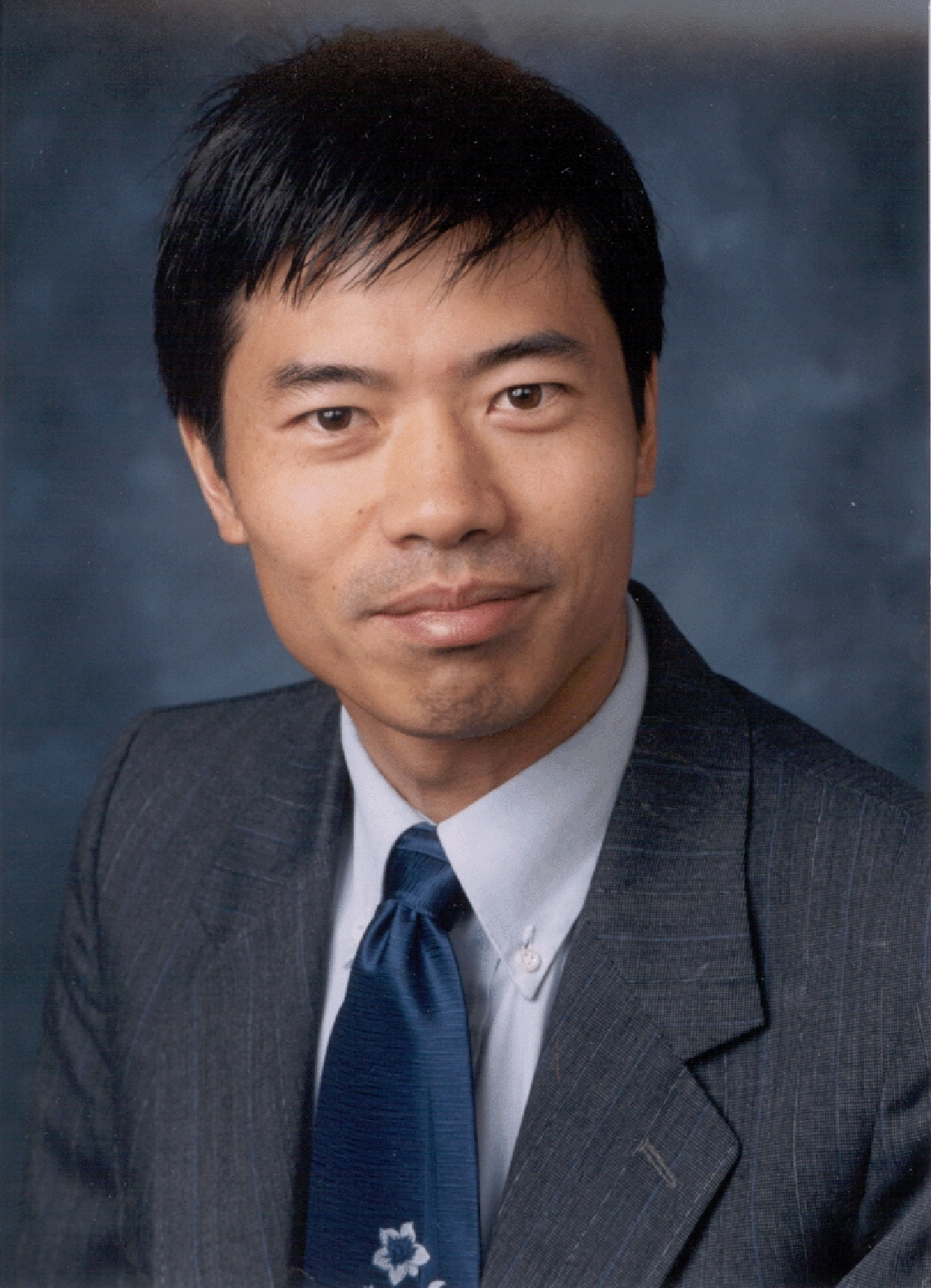}}]{Mengchu Zhou} (Fellow, IEEE) received his Ph. D. degree from Rensselaer Polytechnic Institute, Troy, NY in 1990 and then joined New Jersey Institute of Technology where he has been Distinguished Professor since 2013. His interests are in Petri nets, automation, robotics, big data, Internet of Things, cloud/edge computing, and AI.  He has 1400+ publications including 18 books, 900+ journal papers (700+ in IEEE transactions), 32 patents and 32 book-chapters. He is Fellow of IFAC, AAAS, CAA and NAI.
\end{IEEEbiography}

\vfill

\end{document}